\ifdefined\XeTeXversion\else\pdfoutput=1\fi %% pdfLaTeX (arXiv); skipped under XeTeX/Tectonic
\documentclass[pdflatex,sn-mathphys-num]{sn-jnl}
\usepackage{graphicx}
\usepackage{multirow}
\usepackage{amsmath,amssymb,amsfonts}
\usepackage{amsthm}
\usepackage{mathrsfs}
\usepackage[title]{appendix}
\usepackage{xcolor}
\usepackage{textcomp}
\usepackage{manyfoot}
\usepackage{booktabs}
\usepackage{array}
\usepackage{tabularx}
\usepackage{rotating}
\begin{document}
\title[Sign Language Retrieval in Low-Resource Dictionary Settings]{Studies on Sign Language Retrieval in Low-Resource Dictionary Settings}
\author*[1]{\fnm{Ushnish} \sur{Sarkar}}\email{u.sarkar@vecc.gov.in}
\author[1]{\fnm{Vineet Kumar} \sur{Rakesh}}
\author[1]{\fnm{Bhaswar} \sur{Chattopadhyay}}
\author[1]{\fnm{Tapas} \sur{Samanta}}
\affil*[1]{\orgdiv{Computer \& Informatics Group}, \orgname{Variable Energy Cyclotron Centre}, \orgaddress{\city{Kolkata}, \country{India}}}
\abstract{Most sign languages have no annotated corpus, only small dictionaries with one or a few example videos per sign. This study examines how well a pose encoder trained on a single sign language serves such languages without any target-language training. An encoder trained with metric learning on American Sign Language (ASL-Citizen) is evaluated frozen on six datasets of Turkish, Russian, Indian, Argentinian, Pakistani and Sri Lankan sign languages, with one or four clips per sign enrolled and, where signer identities are available, the query's signer excluded from the gallery. With one enrolled clip, the encoder retrieves the correct sign first in 36.4--80.7 \% of queries, against 16.0--69.6 \% for the best learning-free method (angle-based dynamic time warping) and a six-target mean of 39.3 \% for an ASL-trained OpenHands graph network. In a crossed ablation of the encoder (embedding geometry, aggregation of body and hand embeddings, training objective, and attention between articulators), two choices matter: a per-pair InfoNCE objective outperforms supervised contrastive and cross-entropy training, and blocking attention between body, left hand and right hand inside the encoder raises accuracy by up to 5.1 points. The best configuration reaches 68.0 \% one-shot accuracy (59.0 \% for the reference encoder). On CISLR, an Indian Sign Language dictionary of 4,765 words with one video per word, the encoders find the correct word first for 17.3--19.0 \% of queries, above the 16.8 \% of the original RGB baseline. Over eight seeds with the joint encoder, spherical, Euclidean and hyperbolic embedding spaces differ by at most 1.6 points, except the hyperbolic token-attention head trained with InfoNCE, which is 1.4--2.0 points worse, so the geometry is a minor factor.}
\keywords{sign language retrieval, zero-shot cross-lingual transfer, pose estimation, metric learning, contrastive learning, hyperbolic embeddings}
\maketitle

\section{Introduction}
Sign languages are natural languages with their own lexicon and grammar; more than 300 are used by over 70 million deaf people \cite{ref1}. Annotated video exists at scale for only a few of them. For most sign languages a developer has at best a dictionary with one or a few recorded examples per sign. The Indian Sign Language dictionary of the Indian Sign Language Research and Training Centre, for example, documents 10,000 terms with demonstration videos \cite{ref2}, and a corpus collected from it and a second online dictionary has about 1.5 videos per word \cite{ref3}, whereas INCLUDE, a widely used Indian Sign Language dataset recorded from several signers, covers a few hundred signs \cite{ref4}. Recording a corpus that covers a whole vocabulary, with many signers, repetitions and regional variants of every sign, is slow and expensive, and the effort would have to be repeated for each language. Methods that work from the example videos a dictionary already has, with no training or fine-tuning on the target language, are therefore of direct practical value for under-resourced sign languages.
The simplest useful tool of this kind is a dictionary that is searched by performing a sign in front of a camera. It needs only a function that compares two signing clips \cite{ref5}, and such a search has been deployed for Flemish Sign Language \cite{ref6}. The scientific question behind the tool is one of transfer: signs of different languages are built from shared phonological parameters (handshape, location, movement and orientation) \cite{ref7}, so an encoder trained on one language may capture structure that is useful for others. Which of its properties transfer, and which design choices make it transfer, is less clear. Pose-based encoders are a natural candidate, because keypoints remove appearance and background variation and leave the articulation itself \cite{ref8}; earlier work, surveyed in Section 2, has shown that their embeddings carry information across sign languages, but usually on one or two target languages, in single training runs, and without a controlled comparison of the design choices involved.
Three parts of the problem therefore remain open. \textbf{(i)} How far zero-shot transfer reaches: earlier reports cover one or two languages, control signer overlap only where a dataset's splits happen to, and do not compare with learning-free matching under the same protocol. \textbf{(ii)} Which design choices of the encoder matter: the embedding space, the treatment of body and hands, and the objective have not been varied together with matched controls for this task, and hyperbolic geometry has been used in sign models without matched Euclidean and spherical controls. \textbf{(iii)} How reliable the answers are: earlier results are single runs. The present article addresses these three points with one source language, six unseen target languages and a real dictionary, a crossed ablation of the encoder, and repeated training seeds.

\section{Related Work}
\textbf{Pose-based and cross-lingual sign representations.} Keypoint input removes appearance variation and makes models compact \cite{ref8}. Pose embeddings carry information shared across sign languages: classifier retraining transfers them between languages \cite{ref9}, multilingual pretraining reduces data needs \cite{ref10}, and cross-lingual few-shot recognition has been posed as meta-learning \cite{ref11}. CISLR \cite{ref3} treats Indian Sign Language dictionary recognition as one-shot and matches features of a WLASL-trained RGB I3D network to one prototype per word, with 16.8 \% top-1 accuracy over 4,765 words. Closest to the present setting, Vandendriessche et al. \cite{ref6} use a keypoint PoseFormer trained as a classifier on ASL-Citizen for one-shot dictionary search in Flemish Sign Language, and SignSeek \cite{ref12} pretrains per-articulator encoders on four sign languages and evaluates unseen British Sign Language. The authors' earlier study \cite{ref13} trained a BERT-style pose Transformer with supervised contrastive learning on ASL-Citizen and reported zero-shot Recall@1 of 97.1 \% on LSA64 and 88.6 \% on INCLUDE, with every clip of a dataset queried against all others and without signer control or reference methods. A pose-based reverse dictionary trained on WLASL with SupCon and ArcFace losses reached 3.5 \% Recall@1 on a 226-sign subset of AUTSL with four gallery clips per sign \cite{ref14}.
\textbf{Learning-free retrieval.} Dynamic time warping (DTW) \cite{ref15} of keypoint trajectories has a long history in sign search \cite{ref16} and remains competitive: DTW over inter-segment angles of MediaPipe landmarks retrieves ASL signs from a 1,113-sign lexicon \cite{ref17}, and per-keypoint DTW of the hand trajectories is the best distance-based measure on ASL-Citizen \cite{ref18}. A related one-shot task, spotting a dictionary sign in continuous signing, has been addressed with attention over video features \cite{ref19}.
\textbf{Embedding geometry and objectives.} Hyperbolic spaces embed hierarchies with low distortion \cite{ref20,ref21} and have improved metric learning \cite{ref22}. In sign language, Geo-Sign uses a hyperbolic contrastive regulariser over body-part features for translation \cite{ref23}, and LA-Sign compares manifolds for sign--text alignment, with gains of at most 0.8 points \cite{ref24}. For the objective, supervised contrastive learning (SupCon) \cite{ref25} and InfoNCE \cite{ref26} are standard, and proxy losses \cite{ref27} avoid pair sampling.
\textbf{Position of the present work.} Transfer of ASL-trained pose encoders to unseen languages has thus been shown, and both learned and DTW retrieval are established. Each, however, has been evaluated on one or two target languages, in single runs, and without a controlled comparison of the encoder's design choices. The present study evaluates one source encoder on six unseen languages and a real dictionary, signer-controlled where signer identities are available, places learning-free and learned references under the same trials, and varies geometry, aggregation, objective and articulator attention in one crossed design with repeated seeds.

\section{Aim and Contributions}
\begin{figure}[htbp]
\centering
\includegraphics[width=\textwidth]{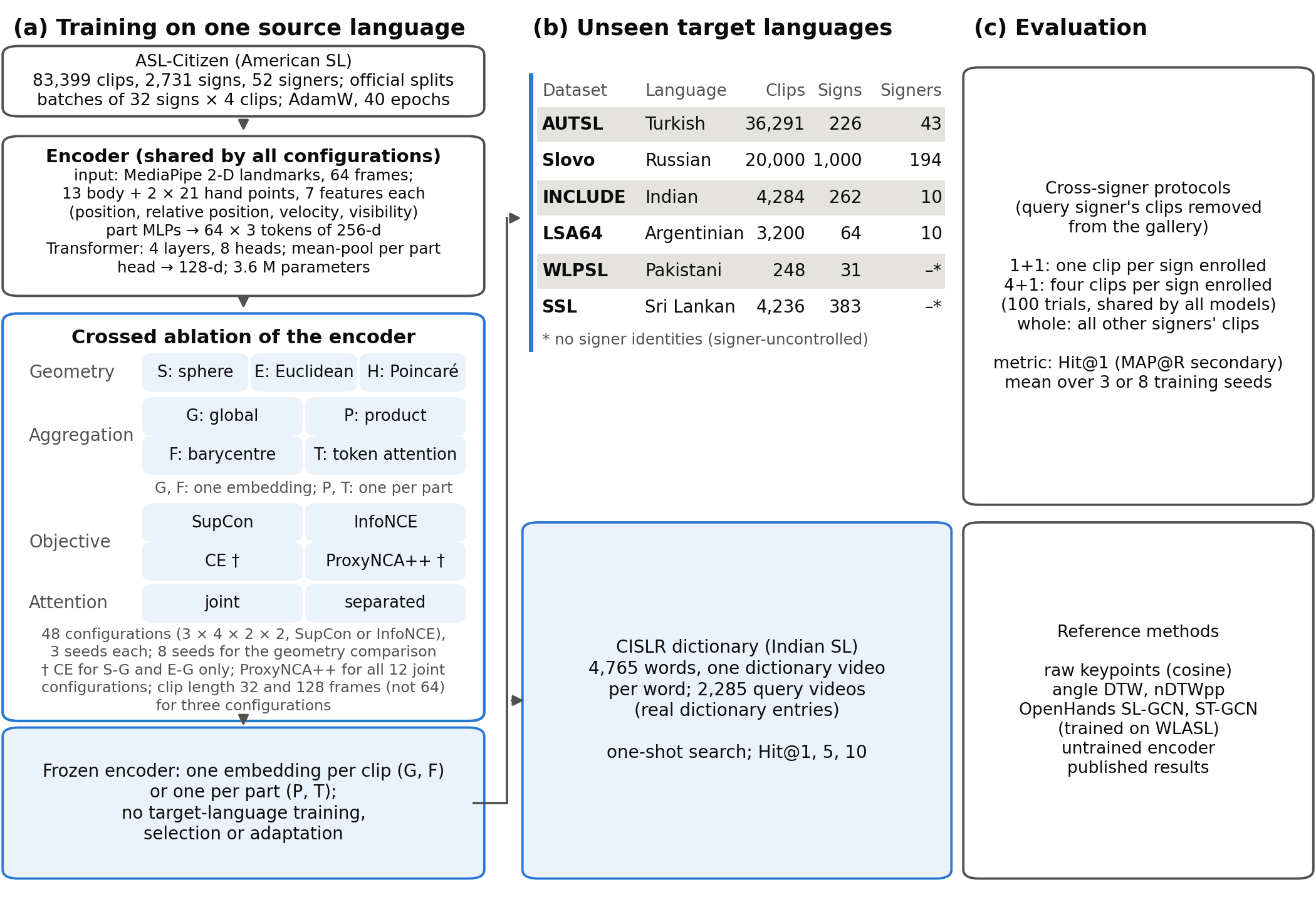}
\caption{Overview of the study. (a) Encoders, all sharing the pose Transformer shown, are trained on ASL-Citizen only, in a crossed grid of embedding geometry (S: unit sphere; E: Euclidean space; H: Poincar\'{e} ball), aggregation of the body and hand embeddings (G: one global embedding; P: one embedding per part, compared by a product similarity; F: the part embeddings merged by a weighted barycentre; T: each part embedding attending over its frame tokens), objective (SupCon or InfoNCE for every configuration; cross-entropy (CE) and ProxyNCA++, marked $\dagger$, for the joint encoder only) and attention between articulators (joint: every token attends to every other; separated: only within its own part), and then frozen. A configuration is named by its geometry and head, e.g. S-G for the global head on the sphere. (b) The frozen encoder embeds clips of six unseen sign languages and of the CISLR Indian Sign Language dictionary. (c) Retrieval is evaluated under cross-signer protocols and compared with learning-free, ASL-trained and published references.}\label{fig:1}
\end{figure}

The aim of this article is to measure how well a pose encoder trained on ASL retrieves signs of unseen sign languages when one or four example clips per sign are available, and to determine which design choices affect this. \textit{Zero-shot} refers to the target language: no data of it are used for training, checkpoint selection or adaptation. The contributions are: (1) an evaluation on six unseen languages, signer-controlled where signer identities are available, against two DTW methods, two learned ASL-trained encoders and untrained references, with repeated training seeds, and a dictionary search over real Indian Sign Language dictionary videos (CISLR); (2) a crossed ablation of geometry (spherical, Euclidean, hyperbolic), aggregation of body and hand embeddings, objective (cross-entropy, SupCon, InfoNCE, ProxyNCA++) and attention between articulators, with up to eight seeds; (3) the finding that the objective and the separation of articulators inside the encoder each change accuracy by several points, whereas the geometry changes it by at most 1.6 points in all but one configuration. Figure~\ref{fig:1} summarises the design of the study.
The remainder of the article is organised as follows. Section 4 describes the datasets, Section 5 the preprocessing of the videos into encoder input, Section 6 the evaluation methodology, Section 7 the model and its training, and Section 8 the variations and ablations that were run. Section 9 reports the results with comparisons to published work, Section 10 discusses them and their limitations, and Section 11 concludes.

\section{Datasets}
\begin{table}[htbp]
\caption{Datasets (clip counts after landmark extraction). LSA64 has 10 signers; subject 10 was recorded in two sessions with disjoint signs, so the protocols use 11 signer identities (IDs). WLPSL and SSL have no signer identities, so their results are signer-uncontrolled.}\label{tab:1}%
\footnotesize\setlength{\tabcolsep}{3pt}
\begin{tabularx}{\linewidth}{@{}>{\raggedright\arraybackslash\hsize=0.971\hsize}X>{\raggedright\arraybackslash\hsize=1.029\hsize}Xcccc@{}}
\toprule
Dataset & Language & Clips & Signs & Signers & Role \\
\midrule
ASL-Citizen & American SL & 83,399 & 2,731 & 52 & source \\
\addlinespace[3pt]
AUTSL & Turkish SL & 36,291 & 226 & 43 & target \\
\addlinespace[3pt]
Slovo & Russian SL & 20,000 & 1,000 & 194 & target \\
\addlinespace[3pt]
INCLUDE & Indian SL & 4,284 & 262 & 10 & target \\
\addlinespace[3pt]
LSA64 & Argentinian SL & 3,200 & 64 & 10 (11 IDs) & target \\
\addlinespace[3pt]
WLPSL & Pakistani SL & 248 & 31 & n/a & target \\
\addlinespace[3pt]
SSL & Sri Lankan SL & 4,236 & 383 & n/a & target \\
\botrule
\end{tabularx}
\end{table}
\textbf{Source dataset.} All encoders proposed in this study are trained on ASL-Citizen \cite{ref28}, a community-sourced dataset of isolated ASL signs recorded by signers in their own homes, with its official signer-independent splits (40,154 training clips from 35 signers; checkpoints are selected on the 10,304-clip validation split). Table~\ref{tab:1} lists the datasets.
\textbf{Target datasets.} The six targets are AUTSL (Turkish) \cite{ref29}, Slovo (Russian; the 400 \textit{no\_event} clips are excluded) \cite{ref30}, INCLUDE (Indian) \cite{ref4}, LSA64 (Argentinian) \cite{ref31}, WLPSL (Pakistani) \cite{ref32} and SSL (Sri Lankan) \cite{ref33}. They span vocabularies from 31 to 1,000 signs and, where signer identities are known, from 10 to 194 signers, and were chosen because they are publicly available isolated-sign corpora of languages unrelated or distantly related to ASL. No target data were used to fit any model, to normalise inputs or to select checkpoints; preliminary target results did, however, inform which further experiments were run, and Section 8 lists them.
\textbf{Dictionary dataset.} CISLR \cite{ref3} was collected from the ISLRTC dictionary and a second online Indian Sign Language dictionary. The present work uses release v1.5-a, whose split sizes equal those of v1.0-a in the original paper: 4,765 words with one dictionary video each, and 2,285 further videos that serve as queries. No CISLR data are used for training or selection. For 422 of its videos (347 dictionary entries, 75 queries) the file header reports 1--12 more frames than can be decoded; these were accepted when an independent decoder (ffprobe) decoded the same number of frames as the landmark extractor, with no other change.
\textbf{Duplicate recordings.} Comparing the extracted landmarks byte for byte shows 42 groups of identical recordings (86 clips) in SSL, 23 of them with different labels, 26 groups in AUTSL (53 clips, each group one sign and one signer in one split) and 1 in Slovo (two clips with different labels); INCLUDE, LSA64 and WLPSL have none. Each AUTSL and Slovo group has a single signer, so the cross-signer protocols already keep these duplicates out of each other's galleries; SSL has no signer identities. Removing every query's duplicates from its gallery changes the reported Hit@1 by $-$0.36 to +0.45 points over the 415 trained models of the independent verification, targets and protocols, so the datasets are kept as published.

\section{Preprocessing}
\begin{figure}[htbp]
\centering
\includegraphics[width=\textwidth]{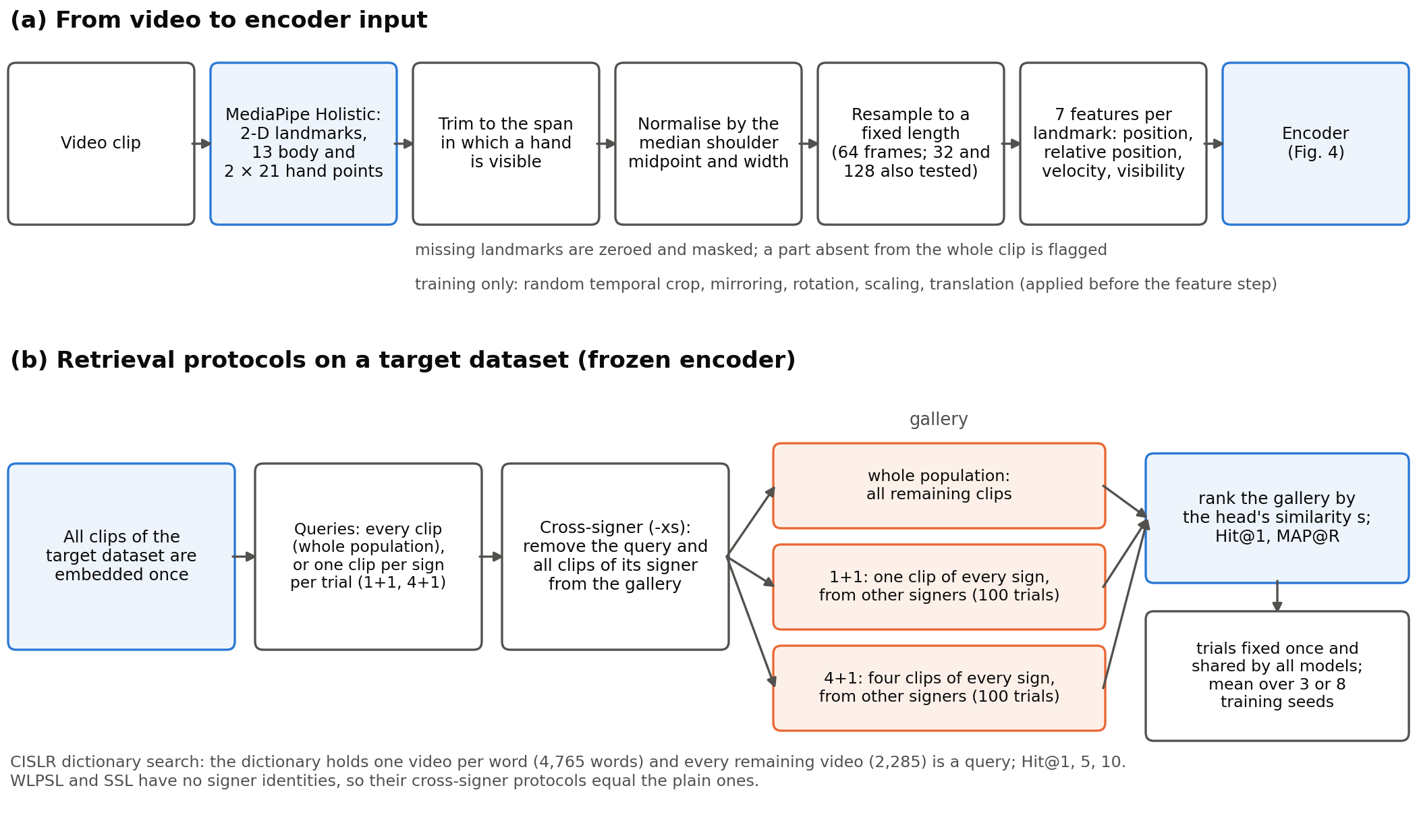}
\caption{Flow charts of the data path. (a) Preprocessing from a video clip to the encoder input: landmark extraction, trimming, normalisation, resampling and per-landmark features; augmentation is applied during training only. (b) The retrieval protocols on a target dataset: in the whole-population protocol every clip is a query against all remaining clips; in the 1+1 and 4+1 protocols each of 100 fixed trials, shared by all models, draws one query and one or four gallery clips per sign; cross-signer variants exclude the query's signer from the gallery.}\label{fig:2}
\end{figure}

\begin{figure}[htbp]
\centering
\includegraphics[width=\textwidth]{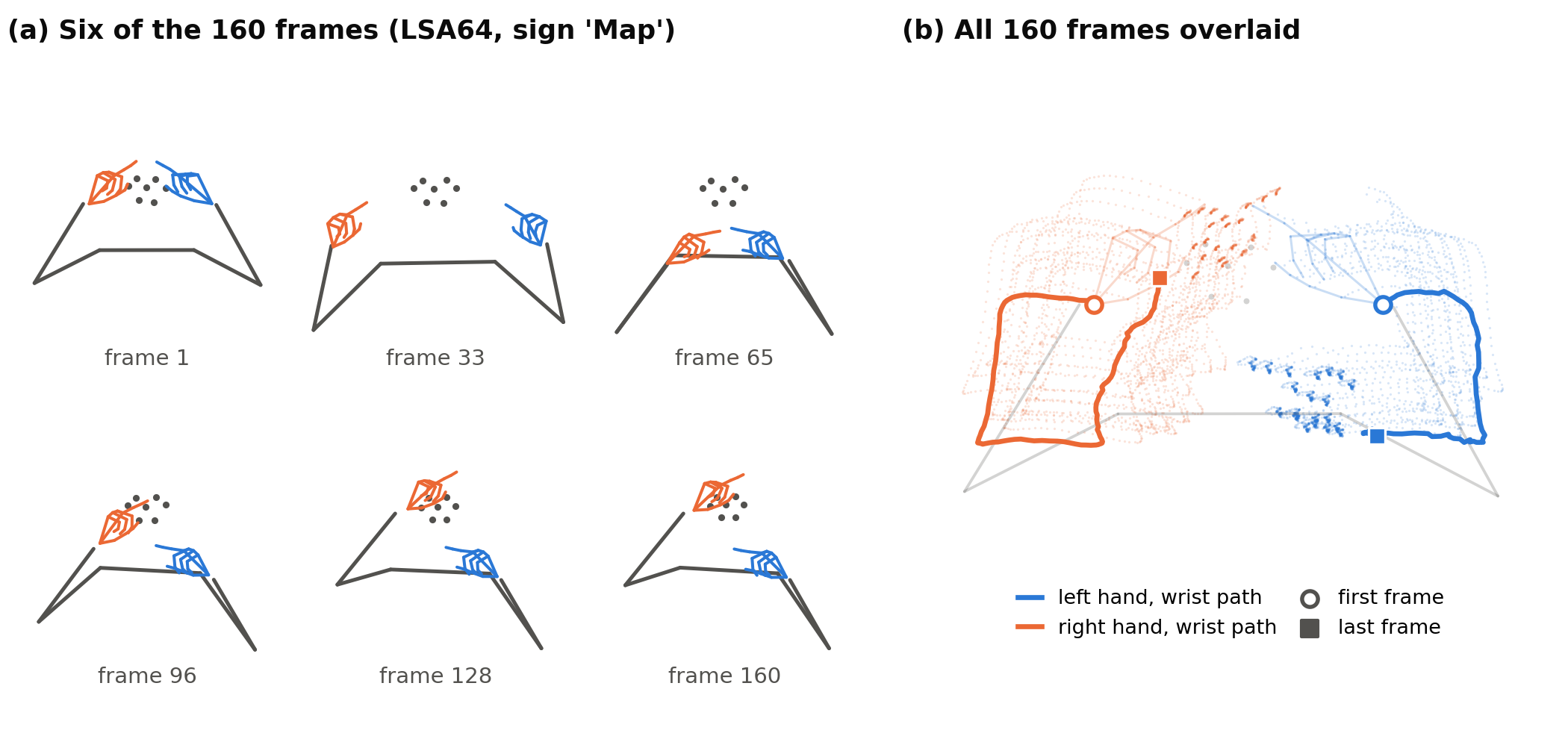}
\caption{Landmarks of one clip as the encoder receives them (LSA64, sign 'Map'), after trimming and normalisation by the shoulder midpoint and width; camera view. (a) Six evenly spaced frames: body (grey; face points as dots), left hand (blue) and right hand (orange), as labelled by MediaPipe. (b) The hand landmarks of all frames (dots) and the path of each wrist from the first frame (circle) to the last (square). The clip is resampled to 64 frames before encoding.}\label{fig:3}
\end{figure}

Figure~\ref{fig:2}a shows the path from a video to the encoder input; the same pinned extractor and the same steps are applied to every dataset, including CISLR. MediaPipe Holistic \cite{ref34} provides 2-D landmarks per frame; 21 points per hand and 13 upper-body points (nose, eyes, ears, mouth corners, shoulders, elbows and wrists) are kept, and the remaining body and face landmarks are discarded. Each clip is trimmed to the span in which at least one hand is detected, which removes the rest position at the start and end of a recording. The landmarks are then normalised by the clip's median shoulder midpoint and shoulder width, so that the position and scale of the signer in the frame do not matter, and the clip is resampled to 64 frames by nearest-frame selection. Sixty-four frames is of the order of the median clip length after trimming (39 frames in ASL-Citizen and 30--93 in the targets, recorded at 10--60 frames per second), so a typical sign is kept near its native rate; the fixed length also makes the input independent of frame rate and signing speed, which differ between datasets. The clip length is varied in the ablation of Section 8. Each landmark is finally described by seven features: its position, its position relative to a part anchor (the wrist for the hands, the shoulder midpoint for the body), its velocity between consecutive frames, and a visibility flag. Landmarks that MediaPipe does not detect are zeroed and masked, and a part that is never visible in a clip is flagged so that the encoder can treat it as absent.
Figure~\ref{fig:3} shows one clip after these steps. During training only, each clip is augmented by a random temporal crop, horizontal mirroring (which also swaps the left and right hands), a rotation of up to $\pm$15$^{\circ}$, scaling by 0.85--1.15 and a translation of up to 0.1 shoulder widths, applied before the feature step so that the velocities stay consistent with the positions. No augmentation and no per-dataset statistics are used at evaluation time.

\section{Evaluation Methodology}
\textbf{Retrieval protocols.} Figure~\ref{fig:2}b shows the protocols. In the \textit{whole-population} protocol every clip of a target dataset is a query and is ranked against all remaining clips. The few-shot protocols simulate a dictionary in sampled episodes: each of 100 trials draws, for every sign that can supply them, one query clip and \textit{k} $\in$ \{1, 4\} gallery clips of that sign, and the query is ranked against the \textit{k} clips of every sign (1+1 and 4+1). In the \textit{cross-signer} variants (suffix -xs), gallery clips of the query's signer are excluded, so the encoder must match different people; this is the regime of a dictionary searched by a new user. Trials depend only on labels and signers, are drawn once with a fixed seed, and are shared by all models, so every model is scored on the same queries and galleries. On CISLR the protocol of the original paper is followed: the one dictionary video of each word forms the gallery and the remaining videos are the queries; a query is correct at rank k when its word is among the k most similar entries.
\textbf{Metrics.} The primary metric is Hit@1 (top-1 accuracy, also called Recall@1), averaged over queries; MAP@R \cite{ref35}, which scores the whole ranking, is reported in the supplementary material. For CISLR, Hit@5 and Hit@10 are also reported because the original paper reports them.
\textbf{Reference methods.} Six reference methods, none trained on ASL-Citizen, are scored on the same trials. Two are learning-free: raw keypoint sequences compared by cosine similarity, and the encoder at random initialisation. Two are published DTW methods: nDTWpp, reproduced from the authors' toolkit \cite{ref18} (the toolkit's name for the nDTWp of that paper), and angle DTW, re-implemented from the paper \cite{ref17} (it reaches 69.6 \% one-shot Hit@1 on LSA64 under signer control, against 72.4 \% reported with two gallery clips). The last are the OpenHands \cite{ref8} SL-GCN \cite{ref36} and ST-GCN \cite{ref37} encoders trained on WLASL \cite{ref38}, used frozen with their own 27-keypoint input and cosine similarity. The DTW methods use the published settings on the 2-D landmarks of Section 5.
\textbf{Repetition.} Every configuration is trained with three seeds, and the joint-encoder configurations with eight, and results are reported as the mean (and standard deviation) over seeds. Because the trials are shared by all models, differences between configurations are read on the same queries and galleries. Full tables for every configuration, protocol and metric are in the supplementary material.

\section{Model}
\textbf{Encoder.} All variants share the encoder of Fig. 4a. The landmarks of each frame are split into three parts (body, left hand, right hand), and each part is turned into one 256-dimensional token by its own two-layer MLP, so a clip becomes 64 $\times$ 3 tokens; learned embeddings mark the part and the frame of each token. A four-layer Transformer \cite{ref39} (eight heads, pre-norm, feed-forward width 1,024, dropout 0.1) processes all tokens together. Finally, the tokens of each part are averaged over the frames in which that part is visible, giving three part vectors h\textsubscript{B}, h\textsubscript{L} and h\textsubscript{R}. The encoder comes in two versions that differ only in the attention mask (Fig. 4b). In the \textbf{joint} encoder every token may attend to every other, so the body vector also carries information about the hands. In the \textbf{separated} encoder a token may attend only to tokens of its own part, so each part vector describes that articulator alone.
\begin{figure}[htbp]
\centering
\includegraphics[width=\textwidth]{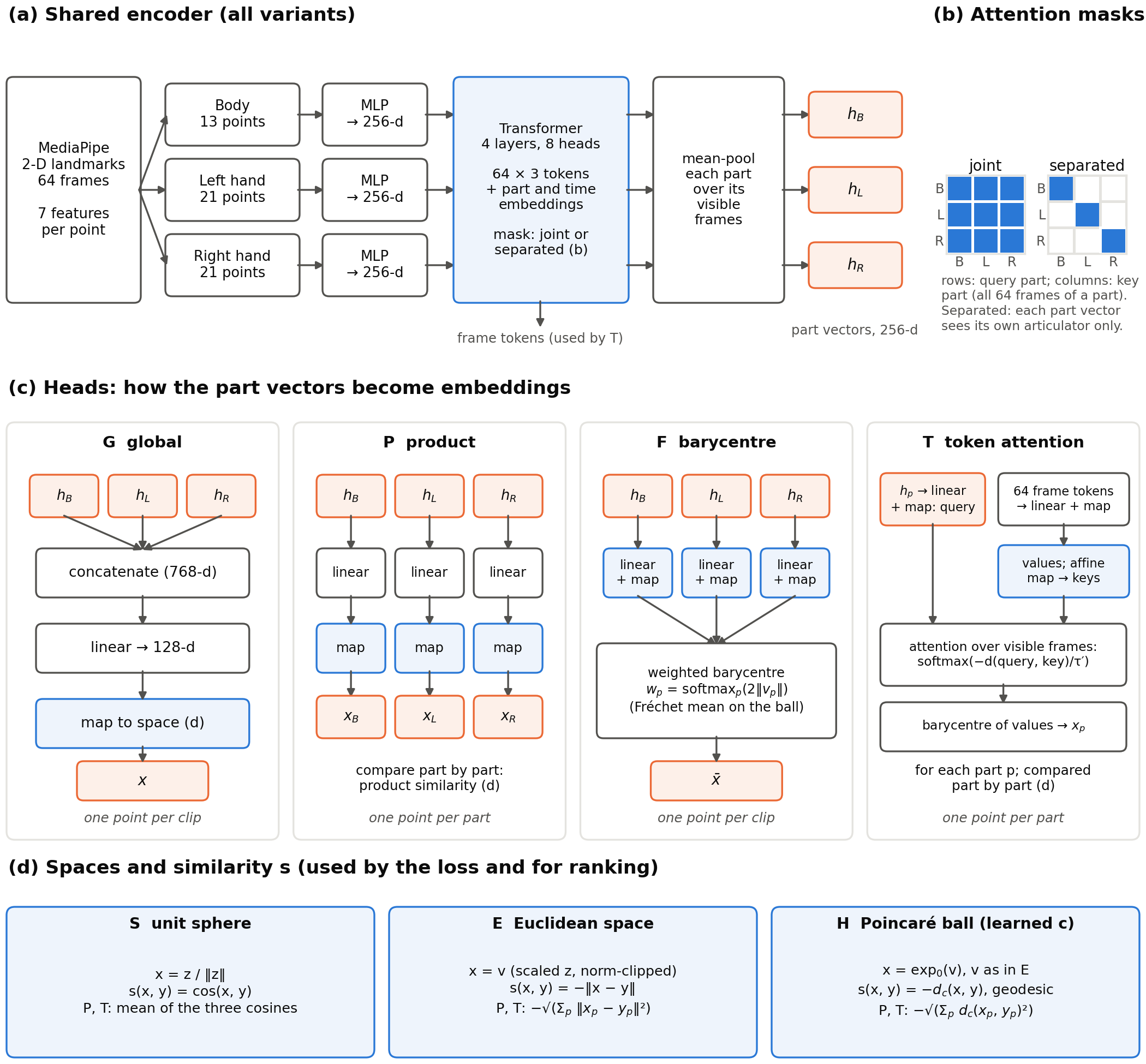}
\caption{Block diagram of the encoder and of every variant. (a) Shared encoder: per-part MLPs turn the landmarks of each frame into 64 $\times$ 3 tokens, a Transformer processes them, and each part is averaged over its visible frames into a part vector (h\textsubscript{B}, h\textsubscript{L}, h\textsubscript{R}). (b) Attention masks over the part blocks (rows: query part, columns: key part): joint, every token attends to every other; separated, only within its own part. (c) The four heads. G concatenates the part vectors into one embedding; P keeps one embedding per part; F merges the per-part embeddings into one point by a weighted barycentre, with v\textsubscript{p} the scaled projection of part p; T lets each part's embedding attend over the frame tokens of that part. (d) The three spaces: how a projection z (or its scaled, norm-clipped version v) becomes a point, and the similarity s used both by the loss and for ranking; heads with one point per part (P, T) combine the parts as shown.}\label{fig:4}
\end{figure}
\textbf{Heads.} A head turns the three part vectors into the embedding used for retrieval (Fig. 4c); every projection is linear and 128-dimensional. \textbf{G} (global) concatenates the three part vectors and projects them to a single embedding. \textbf{P} (product) projects each part separately and keeps three embeddings, which are compared part by part. \textbf{F} (barycentre) also projects each part separately, then merges the three points into one by a weighted barycentre (below). \textbf{T} (token attention), a geometric adaptation of the Geo-Sign token method \cite{ref23}, uses each part's embedding as a query that attends over the 64 frame tokens of the same part, projected into the same space; the attention-weighted barycentre of those tokens becomes that part's embedding, and the three are compared part by part.
\textbf{Spaces.} Each head is placed in one of three spaces (Fig. 4d): the unit sphere with cosine similarity (S); Euclidean space with negative distance (E); or the Poincar\'{e} ball of learned curvature with negative geodesic distance (H) \cite{ref21}, with feature clipping \cite{ref40}. E uses the same learned scale and norm clipping as H, so E versus H isolates curvature. For P and T, the per-part similarities are combined: their mean on the sphere, and the negative root of the summed squared distances in E and H. A configuration is named by its space and head (e.g. S-G, the global head on the sphere), with `separated' added when the encoder uses the separated attention mask. With z the linear projection of a part vector (or of the concatenated part vectors for G), v = clip($\sigma$$\cdot$z) its scaled, norm-clipped version, and c > 0 the learned curvature, the embedding x and the similarity s between two embeddings x and y are:
\begin{equation}
\mathrm{S}:\quad x = z/\|z\|, \qquad s(x, y) = x \cdot y\label{eq:1}
\end{equation}
\begin{equation}
\mathrm{E}:\quad x = v, \qquad s(x, y) = -\|x - y\|\label{eq:2}
\end{equation}
\begin{equation}
\begin{split}\mathrm{H}:\quad x &= \exp_{0}^{c}(v) = \tanh(\sqrt{c}\,\|v\|)\,\frac{v}{\sqrt{c}\,\|v\|}, \\ s(x, y) &= -d_{c}(x, y) = -\frac{2}{\sqrt{c}}\tanh^{-1}\!\big(\sqrt{c}\,\|(-x) \oplus_{c} y\|\big)\end{split}\label{eq:3}
\end{equation}
where $\oplus$\textsubscript{c} is M\"{o}bius addition on the ball \cite{ref21} and the clipping bounds $\surd$c $\|$v$\|$ $\le$ 2.3 \cite{ref40}. For the P and T heads, which keep one point x\textsubscript{p} per part, the similarity is the mean of the three cosines on the sphere and $-$($\Sigma$\textsubscript{p} d(x\textsubscript{p}, y\textsubscript{p})\textsuperscript{2})\textsuperscript{1/2} in E and H.
\textbf{Barycentre head.} For the F head, each part p $\in$ \{body, left, right\} has its own projection v\textsubscript{p} and embedding x\textsubscript{p}. The parts are merged into one point by a weighted barycentre (weighted mean); following Geo-Sign \cite{ref23}, each weight grows with the part's distance from the origin, which on the ball equals 2$\|$v\textsubscript{p}$\|$, and the same rule is used in all three spaces:
\begin{equation}
w_{p} = \frac{\exp(2\|v_{p}\|)}{\sum_{q} \exp(2\|v_{q}\|)}, \qquad \bar{x} = \arg\min_{y} \sum_{p} w_{p}\, d(y, x_{p})^{2}\label{eq:4}
\end{equation}
With d the geodesic distance this is the Fr\'{e}chet (Karcher) mean on the ball \cite{ref41,ref42}, computed from the weighted gyromidpoint by five differentiable Riemannian gradient steps. With the Euclidean distance it is the weighted arithmetic mean, and on the sphere, with the chordal distance, the weighted mean renormalised to unit length. The single point $\bar{x}$ is then compared like a global embedding.
\textbf{Objectives.} With \textit{s}\textsubscript{ij} the head's similarity, $\tau$ a learned temperature, P(i) the other clips of the same sign in the batch and N(i) the clips of other signs, SupCon \cite{ref25} normalises each positive pair over all other clips, whereas InfoNCE \cite{ref26} normalises it over the anchor's negatives only:
\begin{equation}
L_{\mathrm{SupCon}} = -\sum_{i} \frac{1}{|P(i)|} \sum_{p \in P(i)} \log \frac{\exp(s_{ip}/\tau)}{\sum_{a \ne i} \exp(s_{ia}/\tau)}\label{eq:5}
\end{equation}
\begin{equation}
L_{\mathrm{InfoNCE}} = -\sum_{i} \frac{1}{|P(i)|} \sum_{p \in P(i)} \log \frac{\exp(s_{ip}/\tau)}{\exp(s_{ip}/\tau) + \sum_{n \in N(i)} \exp(s_{in}/\tau)}\label{eq:6}
\end{equation}
Two further objectives are used in the ablations. ProxyNCA++ \cite{ref27} replaces the other clips by one learned proxy $\pi$\textsubscript{c} per training sign c, mapped into the head's space like an embedding, and applies a softmax over all proxies, so that every clip is pulled towards its own proxy and pushed from the others; with y(i) the sign of clip i and C the training vocabulary:
\begin{equation}
L_{\mathrm{ProxyNCA++}} = -\sum_{i} \log \frac{\exp(s(x_{i}, \pi_{y(i)})/\tau)}{\sum_{c \in C} \exp(s(x_{i}, \pi_{c})/\tau)}\label{eq:7}
\end{equation}
A plain cross-entropy classifier, the objective of the classifier-trained encoders that have been used frozen for sign retrieval \cite{ref8,ref6}, is trained on the raw projection z\textsubscript{i}, with a weight matrix W and bias b over the |C| = 2,731 training signs:
\begin{equation}
L_{\mathrm{CE}} = -\sum_{i} \log \frac{\exp((W z_{i} + b)_{y(i)})}{\sum_{c \in C} \exp((W z_{i} + b)_{c})}\label{eq:8}
\end{equation}
The classifier is discarded at test time, so retrieval uses the head's similarity s in every case; the four objectives differ only in what the encoder is trained to separate. In the three metric-learning objectives $\tau$ is learned (initialised at 0.1 and bounded to [0.01, 1]); cross-entropy has no temperature. The sums run over the clips of one batch.
\textbf{Training.} Batches hold 32 signs $\times$ 4 clips, preferring different signers. Training uses AdamW \cite{ref43} (learning rate 3 $\times$ 10$^{-4}$, weight decay 0.05) for 40 epochs of 300 steps, with 1,000 warm-up steps and cosine decay; $\tau$ starts at 0.1. The encoder has 3.47 M parameters and the global head 0.10 M (3.57 M in total). The epoch with the best ASL-Citizen validation Hit@1 is kept. No setting was tuned per geometry, per objective or on target data.

\section{Variations and Ablations}
\begin{table}[htbp]
\caption{Design of the study: the factors of the crossed ablation and the further experiments. Run counts are the trained models; the occlusion analysis reuses trained models.}\label{tab:2}%
\footnotesize\setlength{\tabcolsep}{3pt}
\begin{tabularx}{\linewidth}{@{}>{\raggedright\arraybackslash\hsize=0.764\hsize}X>{\raggedright\arraybackslash\hsize=0.866\hsize}X>{\raggedright\arraybackslash\hsize=1.528\hsize}X>{\centering\arraybackslash\hsize=0.842\hsize}Xc@{}}
\toprule
Experiment & Factor or change & Levels & Seeds & Runs \\
\midrule
Main grid & Embedding geometry & S (sphere), E (Euclidean), H (Poincar\'{e} ball) & 0--2; 0--7 for the joint encoder & 264 \\
 & Aggregation (head) & G (global), P (product), F (barycentre), T (token attention) &  &  \\
 & Objective & SupCon, InfoNCE &  &  \\
 & Attention between articulators & joint, separated &  &  \\
\addlinespace[3pt]
Objective ablation & ProxyNCA++ & all 12 joint configurations & 0--2 & 36 \\
 & Cross-entropy classifier & S-G, E-G (joint) & 0--2 & 6 \\
\addlinespace[3pt]
Temperature check & Initial $\tau$ & 0.05, 0.1, 0.2 (S-G, SupCon and InfoNCE) & 0--2 & 12 \\
\addlinespace[3pt]
Dimension sweep & Embedding dimension & G: 24, 96, 384; P: 8, 32, 256 (S, E, H; supplementary) & 0--2 & 54 \\
\addlinespace[3pt]
Clip length & Frames after resampling & 32, 64, 128 (S-G SupCon, S-G InfoNCE, H-G separated InfoNCE) & 0--2 & 18 \\
\addlinespace[3pt]
Articulator occlusion & Parts removed at test time & hands, body, left hand, right hand & 0--2 & -- \\
\botrule
\end{tabularx}
\end{table}
Table~\ref{tab:2} lists the experiments. The \textbf{main grid} crosses the three spaces, the four heads, the two contrastive objectives and the two attention masks: 48 configurations, each trained with three seeds, and the 24 joint-encoder configurations with eight seeds (264 runs). The reference encoder for the transfer comparison is S-G trained with SupCon, chosen before the results were known as the most conventional configuration; the other configurations are reported as the ablation. The geometry comparisons (H versus E, H versus S, E versus S within each head) are the core of the ablation; E and H share the learned scale and the norm clipping, so H versus E isolates the curvature and E versus S the freedom of the norm. Eight seeds were used for the joint configurations because the differences between geometries are of the order of one point, comparable to the seed-to-seed variation of a configuration, whereas the effects of the objective and of articulator separation are several times larger than that variation and are resolved by three seeds.
The \textbf{objective ablation} adds ProxyNCA++ for all twelve joint configurations and a cross-entropy classifier for the spherical and Euclidean global heads, so that pair-based, proxy-based and classification training can be compared on the same encoder. The \textbf{temperature check} retrains S-G with each contrastive loss from three initial temperatures, to test whether the ordering of the losses depends on that setting. The \textbf{dimension sweep} retrains the global head at 24, 96 and 384 dimensions and the product head at 8, 32 and 256, so that a product head with three D-dimensional parts can be matched against a global head of 3D dimensions; its results are in the supplementary material. The \textbf{articulator occlusion} zeroes and masks one part or both hands of every test clip of the trained S-G models, without retraining, to show which articulators carry the transferable information. The \textbf{clip-length ablation} retrains the reference encoder with both contrastive objectives, and the best configuration, with clips resampled to 32 and 128 frames instead of 64. The part-separated encoder, the cross-entropy classifier, the OpenHands references, the eight-seed runs and the clip-length ablation were added after the first three-seed results had been seen and are reported as such; all experiments (173 runs beyond the main grid, 437 in total) follow the same data, schedule and selection rule.

\section{Results}

\subsection{Transfer to unseen sign languages}
\begin{table}[htbp]
\caption{Cross-signer Hit@1 (\%) with one (1+1) and four (4+1) enrolled clips per sign, 100 shared trials: the ASL-Citizen test split (source language, 2,731 signs, signers unseen in training; in-language reference, not part of the mean), the six target languages, and their mean. Encoders of this work: mean over seeds 0--2; S-G is the reference encoder. WLPSL and SSL are signer-uncontrolled. --: the DTW methods were scored on the targets only.}\label{tab:3}%
\scriptsize\setlength{\tabcolsep}{3pt}
\begin{tabularx}{\linewidth}{@{}>{\raggedright\arraybackslash\hsize=1.000\hsize}Xcccccccc@{}}
\toprule
System & \begin{tabular}[b]{@{}c@{}}ASL-Cit.\\test\end{tabular} & AUTSL & Slovo & INCLUDE & LSA64 & WLPSL & SSL & Mean \\
\midrule
\addlinespace[3pt]
\multicolumn{9}{@{}l}{\textit{1+1: one enrolled clip per sign}} \\
Raw keypoints, cosine & 5.3 & 10.5 & 7.0 & 6.6 & 51.1 & 23.5 & 17.0 & 19.3 \\
nDTWpp \cite{ref18} & -- & 8.8 & 5.3 & 13.0 & 49.6 & 30.4 & 17.1 & 20.7 \\
Angle DTW \cite{ref17} & -- & 21.1 & 16.0 & 21.5 & 69.6 & 58.8 & 29.3 & 36.0 \\
Untrained encoder & 4.2 & 10.7 & 6.7 & 13.5 & 45.2 & 38.2 & 23.7 & 23.0 \\
OpenHands \cite{ref8} SL-GCN \cite{ref36}, WLASL \cite{ref38} & 11.3 & 19.1 & 13.8 & 35.3 & 64.3 & 70.6 & 32.8 & 39.3 \\
OpenHands \cite{ref8} ST-GCN \cite{ref37}, WLASL \cite{ref38} & 7.5 & 18.0 & 11.0 & 30.7 & 55.9 & 59.5 & 26.9 & 33.7 \\
\addlinespace[2pt]
\multicolumn{9}{@{}l}{\textit{This work}} \\
\hspace{0.8em}S-G, SupCon & 40.3 & 51.8 & 36.4 & 59.9 & 80.7 & 74.9 & 50.5 & 59.0 \\
\hspace{0.8em}S-G, InfoNCE & 43.9 & 54.3 & 40.6 & 63.5 & 82.4 & 79.0 & 52.9 & 62.1 \\
\hspace{0.8em}\textbf{H-G separated, InfoNCE} & \textbf{47.3} & \textbf{60.0} & \textbf{47.4} & \textbf{70.7} & \textbf{87.7} & \textbf{85.0} & \textbf{57.0} & \textbf{68.0} \\
\addlinespace[3pt]
\multicolumn{9}{@{}l}{\textit{4+1: four enrolled clips per sign}} \\
Raw keypoints, cosine & 9.6 & 16.7 & 12.6 & 10.6 & 68.7 & 38.9 & 21.5 & 28.2 \\
nDTWpp \cite{ref18} & -- & 13.8 & 9.2 & 20.1 & 65.5 & 45.5 & 22.7 & 29.5 \\
Angle DTW \cite{ref17} & -- & 34.5 & 26.9 & 34.3 & 86.0 & 75.5 & 40.1 & 49.6 \\
Untrained encoder & 7.2 & 17.3 & 11.3 & 20.0 & 63.0 & 49.3 & 29.6 & 31.8 \\
OpenHands \cite{ref8} SL-GCN \cite{ref36}, WLASL \cite{ref38} & 19.2 & 29.4 & 22.8 & 48.6 & 79.9 & 81.6 & 44.7 & 51.2 \\
OpenHands \cite{ref8} ST-GCN \cite{ref37}, WLASL \cite{ref38} & 12.9 & 28.9 & 18.2 & 42.7 & 72.1 & 73.9 & 36.1 & 45.3 \\
\addlinespace[2pt]
\multicolumn{9}{@{}l}{\textit{This work}} \\
\hspace{0.8em}S-G, SupCon & 52.7 & 67.5 & 48.3 & 72.2 & 90.0 & 85.3 & 64.0 & 71.2 \\
\hspace{0.8em}S-G, InfoNCE & 57.1 & 70.7 & 53.3 & 76.3 & 91.7 & 88.6 & 66.4 & 74.5 \\
\hspace{0.8em}\textbf{H-G separated, InfoNCE} & \textbf{60.9} & \textbf{75.3} & \textbf{60.1} & \textbf{81.7} & \textbf{94.8} & \textbf{94.1} & \textbf{70.0} & \textbf{79.3} \\
\botrule
\end{tabularx}
\end{table}
\begin{figure}[htbp]
\centering
\includegraphics[width=\textwidth]{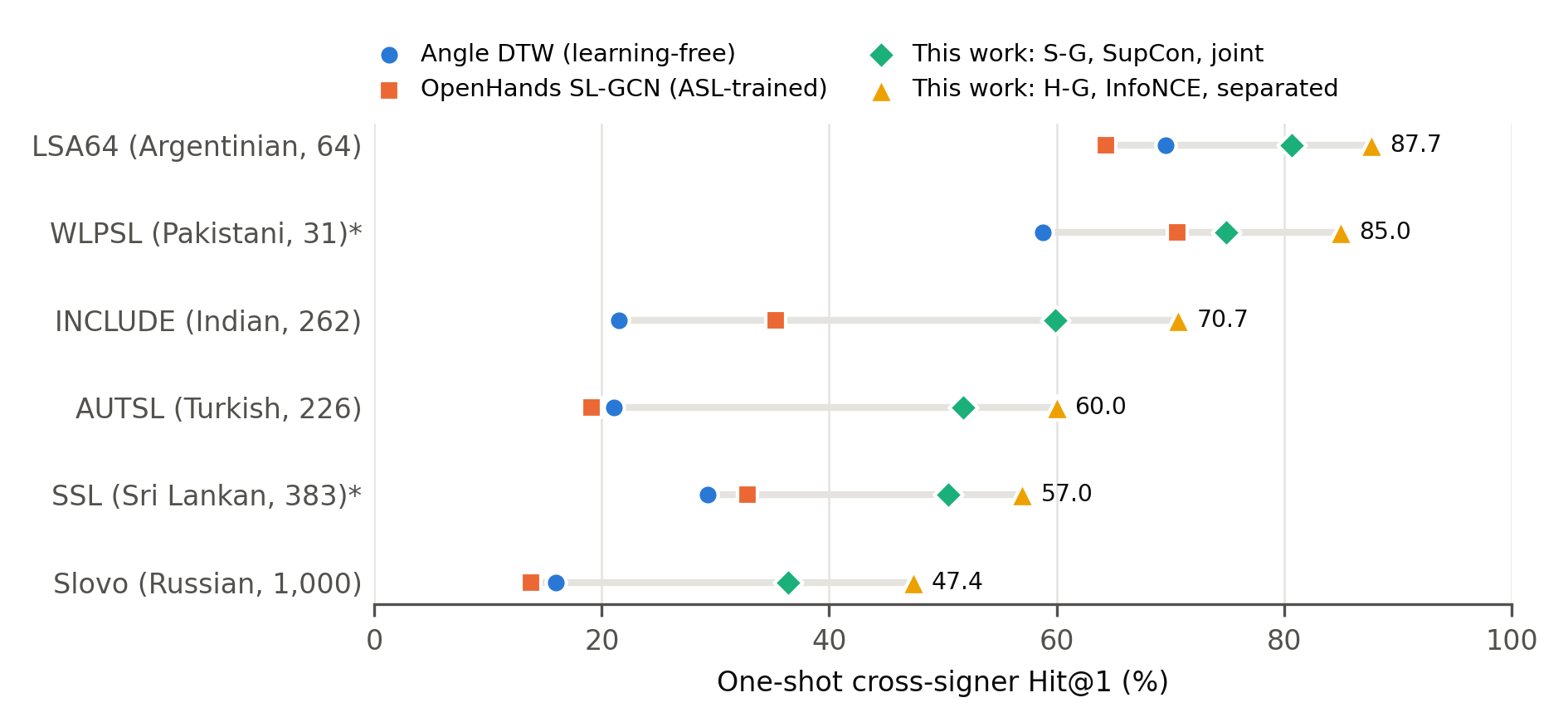}
\caption{One-shot cross-signer Hit@1 per target language (number of signs in parentheses), ordered by the reference encoder (S-G, SupCon); values as in Table~\ref{tab:3}, the encoders of this work averaged over three seeds. * signer-uncontrolled (no signer identities).}\label{fig:5}
\end{figure}
Table~\ref{tab:3} and Figure~\ref{fig:5} show the central result. With a single enrolled clip per sign, the reference encoder (S-G, SupCon) reaches 36.4--80.7 \% Hit@1 across the six languages, against 16.0--69.6 \% for angle DTW, the strongest learning-free method, a mean gain of +23.0 points. It is better than every reference on all six languages under both protocols. The ASL-trained OpenHands graph networks transfer much less well (six-target mean 39.3 \% for SL-GCN), which shows that the gain is not simply a property of any ASL-trained model; they were trained as WLASL classifiers on a different keypoint set, so this compares available encoders rather than objectives. The gap to the references is smallest on WLPSL and LSA64, whose vocabularies are smallest and where learning-free matching is already strong. The best configuration of Section 9.2 raises the six-target mean from 59.0 to 68.0 \% (4+1: 71.2 to 79.3 \%). The first column gives the in-language reference: on the ASL-Citizen test split, whose signers are unseen in training, S-G reaches 40.3 \% one-shot Hit@1 and the best configuration 47.3 \%, against 5.3 \% for raw keypoints. These values are lower than the target means because the ASL vocabulary is far larger (2,731 signs against 31--1,000), so they are not comparable with the cross-lingual columns in absolute terms; they show the level the encoder reaches on its own language under the same protocol.

\subsection{Ablation of the encoder design}
\begin{figure}[htbp]
\centering
\includegraphics[width=\textwidth]{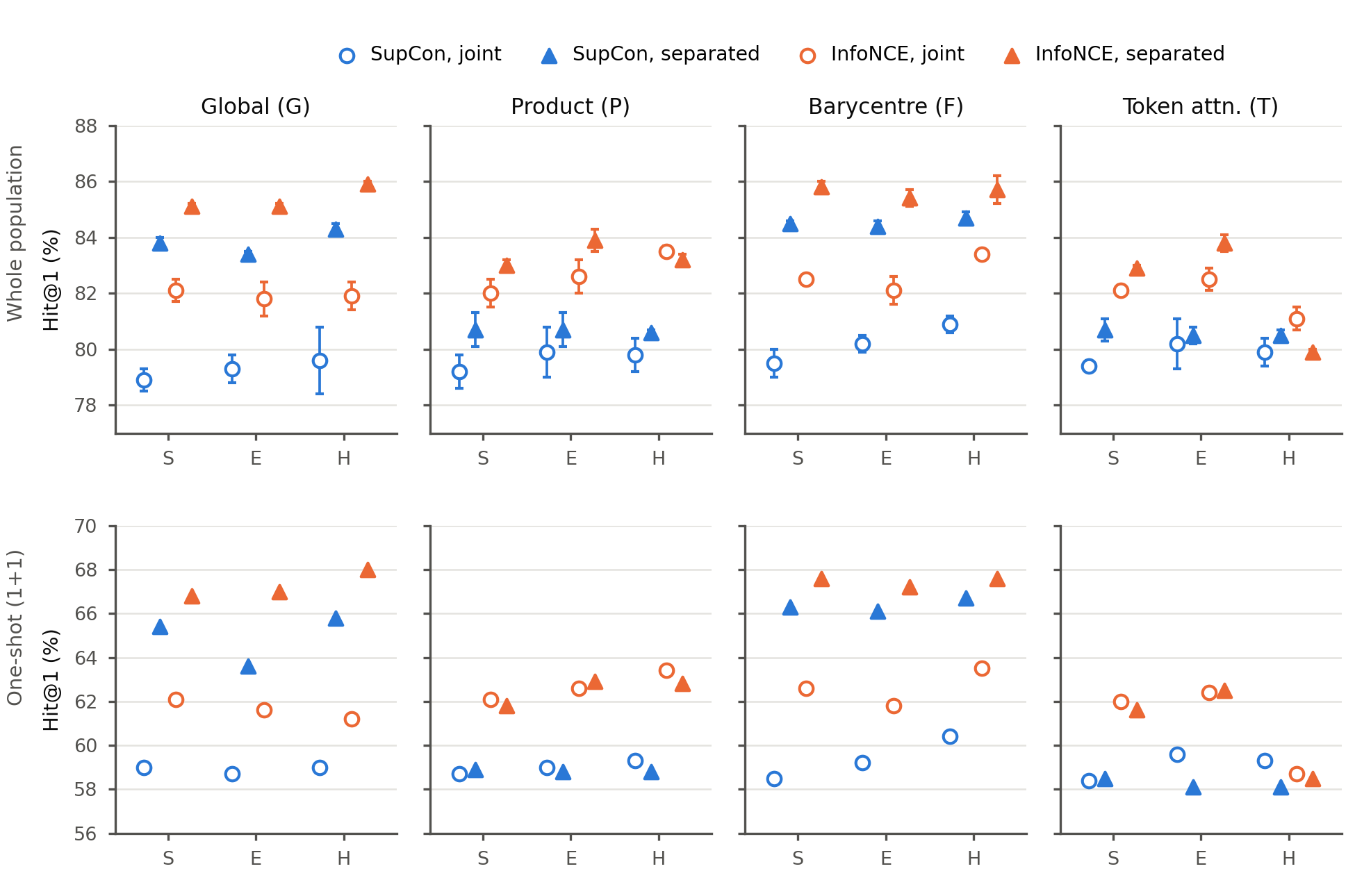}
\caption{Ablation of geometry, aggregation, objective and attention between articulators: six-target mean cross-signer Hit@1 (\%) on the whole population (top row; whiskers: SD over seeds 0--2) and with one enrolled clip (bottom row), one panel per aggregation head, geometry on the horizontal axis. Colour: objective; marker: encoder (open circle, joint attention; filled triangle, attention only within body, left hand and right hand). The values are tabulated in the supplementary material.}\label{fig:6}
\end{figure}
Differences quoted in the text are computed from the unrounded means and rounded once, so they can differ by 0.1 point from differences between the rounded table values. \textbf{Separating the articulators.} Blocking attention between body and hands is the largest single change (Figure~\ref{fig:6}). For heads that pool the parts into one embedding (global and barycentre), it changes whole-population Hit@1 by +3.8 to +5.1 points with SupCon and +2.2 to +4.0 with InfoNCE, on every target; for heads that keep three part embeddings to the end (product and token attention) the change is +0.3 to +1.5 and $-$1.1 to +1.3. Separation inside the encoder therefore helps, but keeping the parts separate in the retrieval space does not: with the separated encoder the product and token heads are 1.3--6.0 points below the corresponding global head. Retraining the joint S-G encoder with the final code reproduced the stored runs to within 0.6 points, so the gain is not a code artefact.
\textbf{Geometry.} Within each aggregation and encoder, the three geometries differ by at most 1.5 points, except the hyperbolic token-attention head with InfoNCE and the separated encoder, which is several points below the other two. With eight seeds (joint encoder, SupCon), the twelve configurations span 78.9--80.6 \%; no two geometries of the same head differ by more than 1.1 points, the Euclidean and hyperbolic multi-part heads are slightly above their spherical counterparts, and hyperbolic and Euclidean heads, which differ only in curvature, are within 0.5 points of each other. With InfoNCE, hyperbolic space is about one point ahead of its Euclidean control for the product and barycentre heads (+1.1, +1.0), and the hyperbolic token-attention head is 1.4--2.0 points below the other two geometries on all six targets. The geometry is therefore a minor factor for cross-lingual transfer, and curvature by itself gains at most about one point, for two heads under one objective.
\begin{figure}[htbp]
\centering
\includegraphics[width=\textwidth]{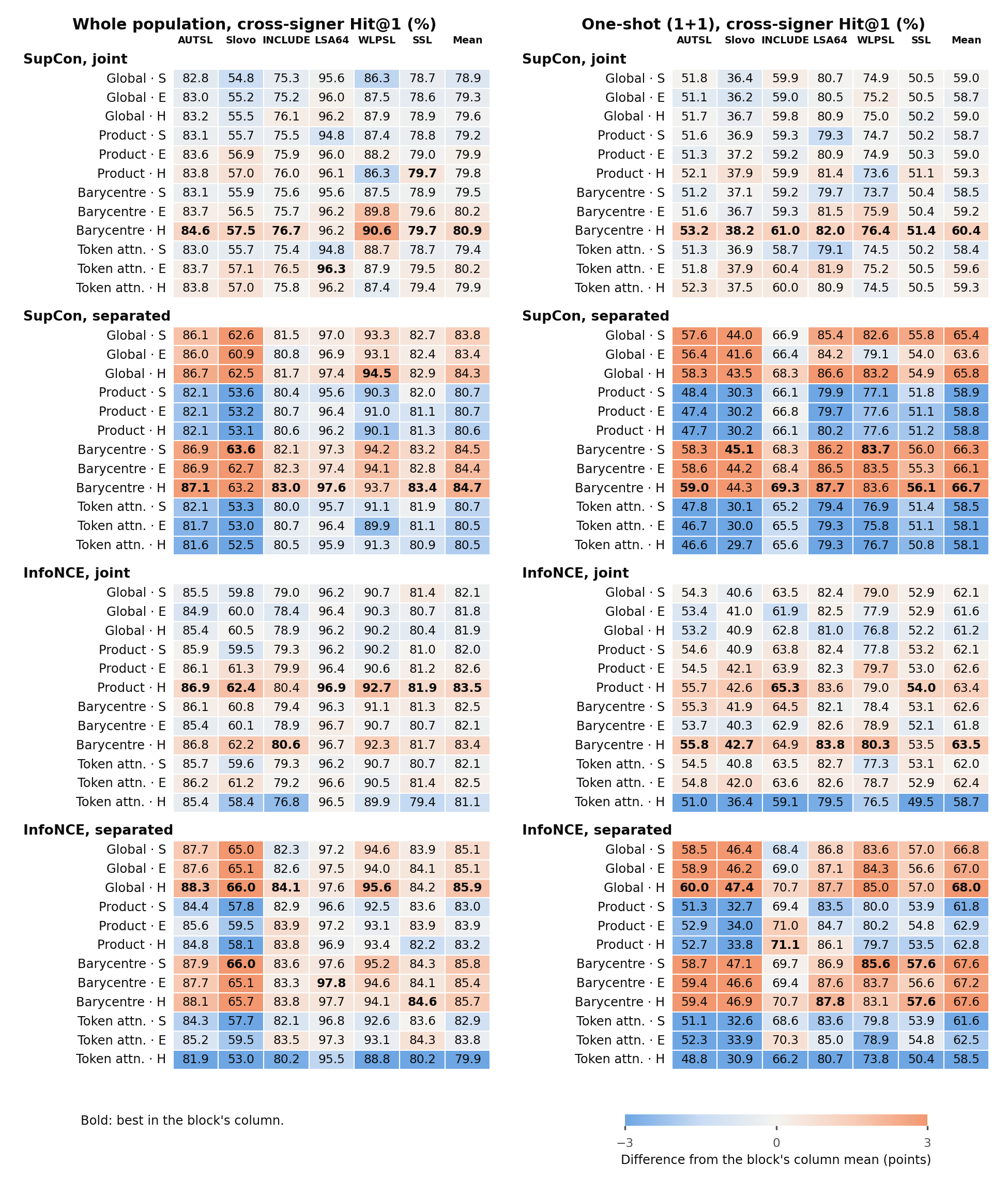}
\caption{Figure~\ref{fig:6} per target. Cross-signer Hit@1 (\%) on the whole population (left) and with one enrolled clip (right), mean over seeds 0--2 (one-shot: also over 100 trials), for every geometry (S, E, H) and aggregation head, in four blocks of objective and encoder; the Mean column reproduces Figure~\ref{fig:6}. The colour of a cell is its difference from the mean of its column within the block (blue below, orange above, saturated at three points), so colours compare heads and geometries on one target and are not comparable between targets or blocks; the best value in each column of a block is bold. The complete tables are in the supplementary material.}\label{fig:7}
\end{figure}
Figure~\ref{fig:7} gives the ablation per target; its mean columns reproduce Figure~\ref{fig:6}. The gain from separating the articulators is positive on every target for every global and barycentre configuration: on the whole population it ranges from +0.9 (LSA64) to +7.8 points (Slovo), and with one enrolled clip from +2.8 to +10.0 points. The three geometries differ more on single targets than in the means: within a head, encoder and objective they span up to 3.1 points on one target (WLPSL) on the whole population and up to 5.5 points with one enrolled clip. The same geometry leads on all six targets in 5 of the 16 objective--encoder--head cells on the whole population (hyperbolic in 4, Euclidean in 1, spherical in 0) and in 3 with one enrolled clip (hyperbolic in 3, Euclidean in 0, spherical in 0), by margins within the spread just given; the eight-seed means of Figure~\ref{fig:6}, not single-target differences, are the basis for judging the geometries.
\begin{figure}[htbp]
\centering
\includegraphics[width=\textwidth]{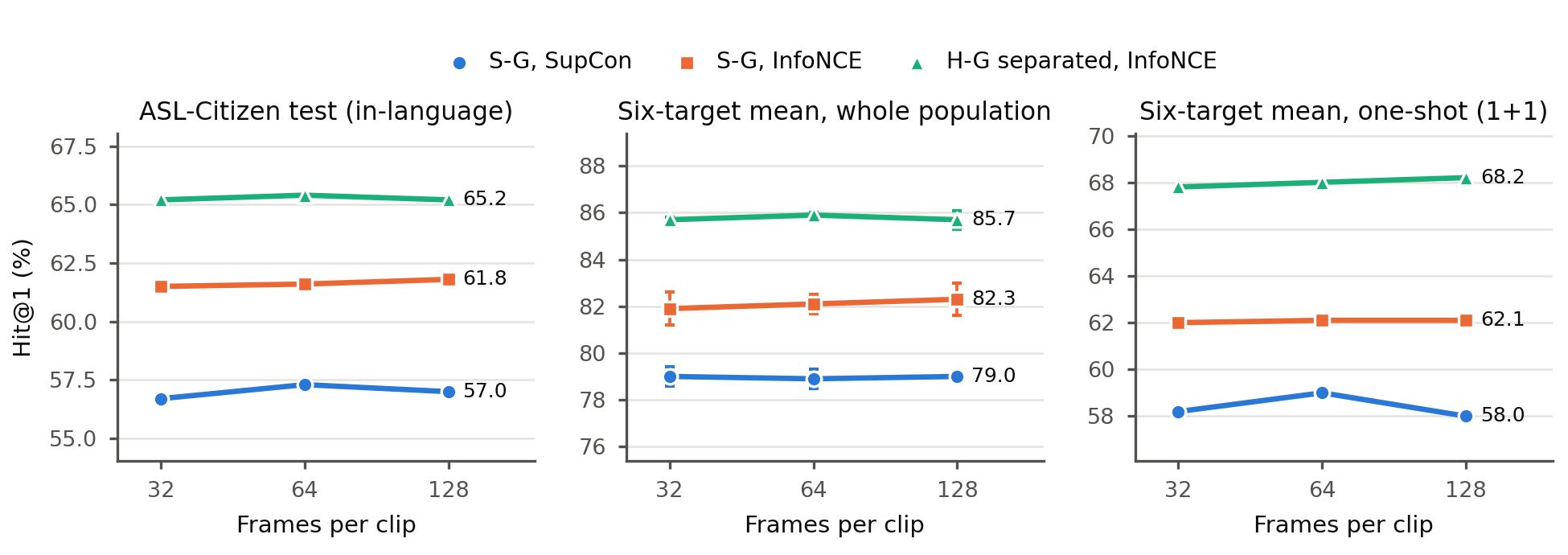}
\caption{Clip length: cross-signer Hit@1 (\%) for clips resampled to 32, 64 (main grid) and 128 frames, for the reference encoder with both contrastive objectives and the best configuration: ASL-Citizen test split (in-language), six-target mean on the whole population (whiskers: SD over seeds 0--2) and six-target one-shot mean. Every panel spans 14 points. The values, per target, are tabulated in the supplementary material.}\label{fig:8}
\end{figure}
\textbf{Clip length.} Figure~\ref{fig:8} varies the number of frames to which every clip is resampled. Relative to 64 frames, 32 frames change the six-target whole-population mean by $-$0.2 to +0.1 points and 128 frames by $-$0.2 to +0.2 points across the three configurations. On the one-shot protocol the corresponding changes are $-$0.8 to $-$0.1 points for 32 frames and $-$1.0 to +0.2 points for 128 frames. Neither length is more than half a point above 64 frames in any configuration, so 32 or 128 frames did not improve the selected endpoints and the conclusions of the ablation do not depend on the choice. With three seeds this does not show that finer temporal information is absent or unusable; the largest one-shot change, a decline at 128 frames, is close to one point. 

\subsection{Training objective}
\begin{figure}[htbp]
\centering
\includegraphics[width=\textwidth]{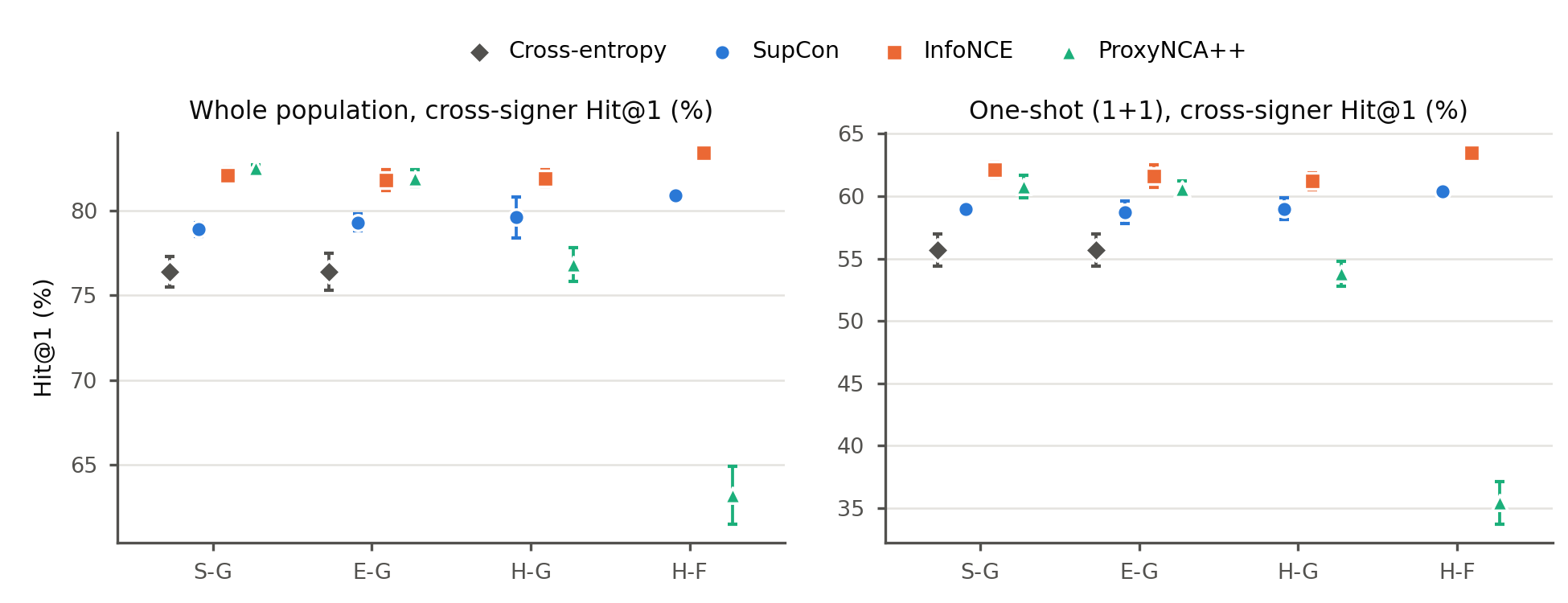}
\caption{Training objective: six-target mean cross-signer Hit@1 (\%) on the whole population (left) and with one enrolled clip (right), joint encoder, for the spherical, Euclidean and hyperbolic global heads and the hyperbolic barycentre head (whiskers: SD over seeds 0--2). Cross-entropy was trained for the spherical and Euclidean global heads only; ProxyNCA++ for all twelve joint configurations, of which the four heads discussed in the text are shown. The values are tabulated in the supplementary material.}\label{fig:9}
\end{figure}
Cross-entropy is the weakest objective and InfoNCE the strongest on the one-shot protocol for the four heads compared in Figure~\ref{fig:9}; on the whole population ProxyNCA++ is within half a point of InfoNCE for the spherical and Euclidean global heads. For the spherical global head, cross-entropy gives 76.4 $\pm$ 0.9 \%, SupCon 78.9 $\pm$ 0.4 \% and InfoNCE 82.1 $\pm$ 0.4 \%; the one-shot ordering is the same. On the whole population InfoNCE beats SupCon in all twelve geometry--aggregation configurations of Figure~\ref{fig:6} with the joint encoder and in eleven of twelve with the separated encoder; with one enrolled clip it does so in eleven of twelve joint configurations (not for the hyperbolic token-attention head, 59.3 against 58.7 \%), and in twelve of twelve separated ones, and for the spherical global head the gap persists when each loss starts from its best temperature. ProxyNCA++ fails for the hyperbolic global and barycentre heads (the hyperbolic product and token-attention heads stay at the SupCon level), most severely for the barycentre head, which also underfits the training data: an optimisation failure of proxies on the ball rather than evidence about the geometry. Objectives that act directly on the embedding (SupCon, InfoNCE, ProxyNCA++) thus transfer better than a discarded classifier; among them InfoNCE is the best on the one-shot protocol, and ProxyNCA++ matches it on the whole population for the heads on which it trains stably.
\begin{figure}[htbp]
\centering
\includegraphics[width=\textwidth]{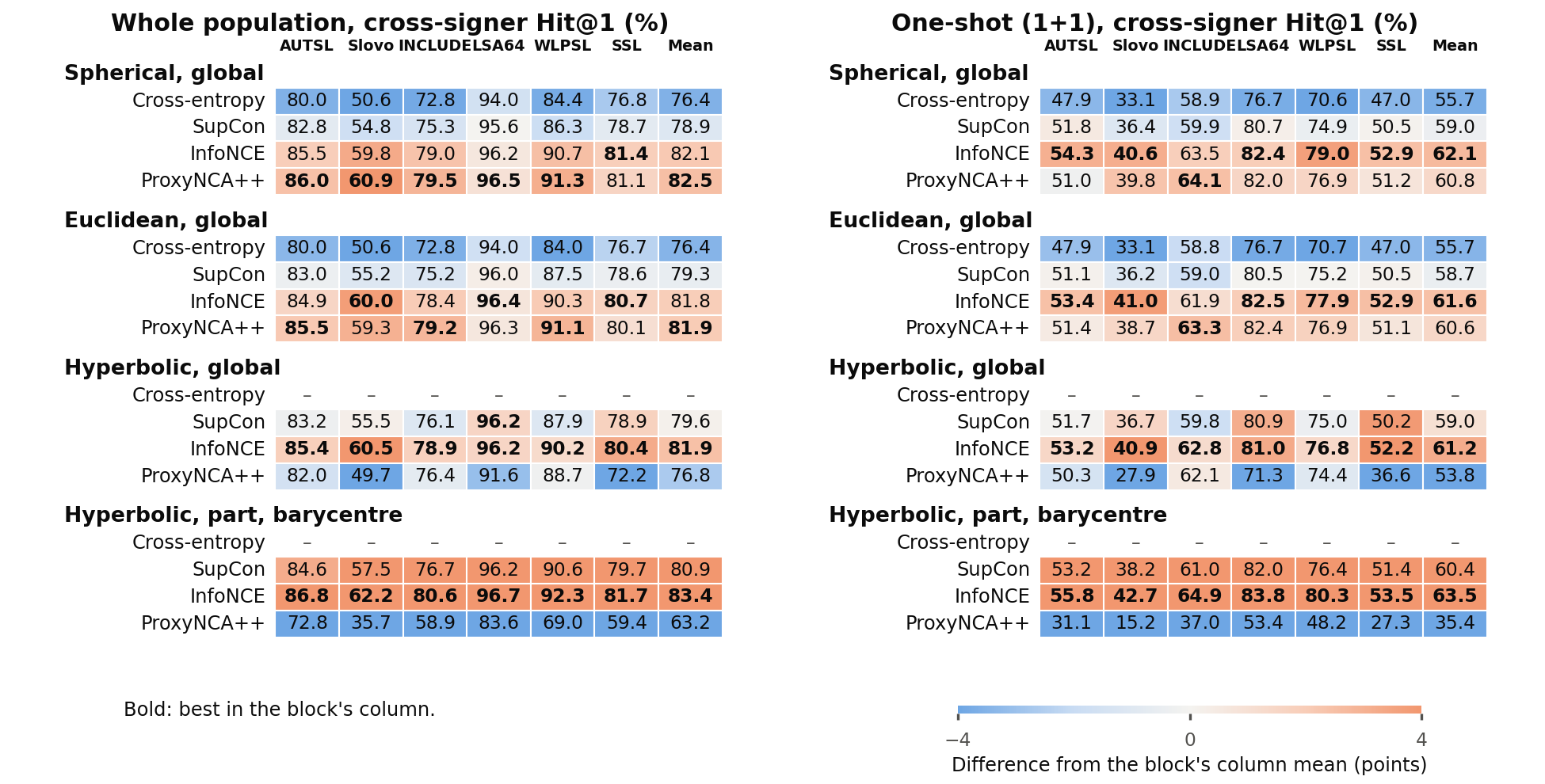}
\caption{Figure~\ref{fig:9} per target. Cross-signer Hit@1 (\%) by training objective for the four heads of the joint encoder, on the whole population (left) and with one enrolled clip (right), mean over seeds 0--2 (one-shot: also over 100 trials); the Mean column reproduces Figure~\ref{fig:9}. Colour: difference from the column mean within the head's block (saturated at four points); bold: best objective for that head and target; a dash marks the heads for which cross-entropy was not trained. The complete table is in the supplementary material.}\label{fig:10}
\end{figure}
Figure~\ref{fig:10} gives the objective comparison per target; its mean columns reproduce Figure~\ref{fig:9}. The ordering in the means holds target by target: cross-entropy is below every contrastive objective on all six targets for both heads on which it was trained, on both protocols, and ProxyNCA++ is below InfoNCE on all six targets for the hyperbolic global and barycentre heads, so its failure there is not driven by one corpus.

\subsection{Which articulators carry the signal}
Removing parts at test time (spherical global head, joint encoder) lowers six-target Hit@1 from 78.9 \% to 30.2 \% without hands, 72.9 \% without the body, 63.5 \% without the left hand and 50.7 \% without the right hand. The hands carry most of the transferable information, the right hand (dominant for most signers) most of all, which is consistent with the benefit of encoding them without interference from the body.

\subsection{Comparison with published results}
\begin{table}[htbp]
\caption{Comparison with published results under their protocols. AUTSL: for each of the 226 signs, four training clips form the gallery and one test clip is the query (train and test signers are disjoint); 100 draws shared by all models; this work: mean $\pm$ SD over three seeds. LSA64 and INCLUDE: every clip queried against all others, same-signer matches allowed. Recall@k = Hit@k (\%); mAP $\times$ 100.}\label{tab:4}%
\footnotesize\setlength{\tabcolsep}{3pt}
\begin{tabularx}{\linewidth}{@{}>{\raggedright\arraybackslash\hsize=1.034\hsize}X>{\raggedright\arraybackslash\hsize=0.966\hsize}Xccc@{}}
\toprule
Protocol & System & R@1 & R@10 & mAP \\
\midrule
AUTSL, 4 gallery clips + 1 query per sign \cite{ref14} & Transformer, WLASL \cite{ref38}, published & 3.5 & 29.2 & 9.1 \\
 & OpenHands \cite{ref8} SL-GCN \cite{ref36}, WLASL \cite{ref38} & 28.7 & 63.0 & 21.6 \\
 & OpenHands \cite{ref8} ST-GCN \cite{ref37}, WLASL \cite{ref38} & 28.9 & 65.8 & 20.7 \\
 & Untrained encoder & 17.8 & 52.3 & 13.7 \\
\addlinespace[2pt]
 & \textit{This work} &  &  &  \\
 & \hspace{0.8em}S-G, SupCon & 66.1 $\pm$ 0.8 & 92.1 $\pm$ 0.5 & 55.4 $\pm$ 0.8 \\
 & \hspace{0.8em}S-G, InfoNCE & 69.4 $\pm$ 0.5 & 93.2 $\pm$ 0.1 & 57.8 $\pm$ 0.5 \\
 & \hspace{0.8em}S-G separated, SupCon & 71.2 $\pm$ 0.3 & 94.7 $\pm$ 0.1 & 60.9 $\pm$ 0.4 \\
 & \hspace{0.8em}H-G separated, InfoNCE & 74.0 $\pm$ 0.5 & 95.6 $\pm$ 0.3 & 63.3 $\pm$ 0.6 \\
\addlinespace[3pt]
LSA64, all clips, same signer allowed & Sarkar et al. \cite{ref13}, published & 97.1 & 99.4 & -- \\
\addlinespace[2pt]
 & \textit{This work} &  &  &  \\
 & \hspace{0.8em}S-G, SupCon & 99.1 & 99.7 & -- \\
 & \hspace{0.8em}H-G separated, InfoNCE & 99.6 & 99.9 & -- \\
\addlinespace[3pt]
INCLUDE, all clips, same signer allowed & Sarkar et al. \cite{ref13}, published & 88.6 & 97.6 & -- \\
\addlinespace[2pt]
 & \textit{This work} &  &  &  \\
 & \hspace{0.8em}S-G, SupCon & 96.1 & 99.1 & -- \\
 & \hspace{0.8em}H-G separated, InfoNCE & 97.6 & 99.4 & -- \\
\botrule
\end{tabularx}
\end{table}
The only published zero-shot AUTSL result found for an encoder trained on another sign language is the reverse dictionary of Orazumbekov et al. \cite{ref14}, trained on WLASL. Their subset of AUTSL is not published, so it is redrawn 100 times (Table~\ref{tab:4}). Under this protocol the joint S-G encoder reaches 66.1 $\pm$ 0.8 \% Recall@1 and the best separated encoder 74.0 $\pm$ 0.5 \%, against 28.7 \% for the WLASL-trained OpenHands SL-GCN and the published 3.5 \%. The published value is below the untrained encoder of this work (17.8 \%), so their sample, preprocessing or training must differ and the comparison is indicative only; the like-for-like comparison with WLASL-trained models is the OpenHands rows. Under the signer-uncontrolled protocol of the authors' earlier study, the present encoders are higher on both LSA64 and INCLUDE. Table~\ref{tab:3} and Figure~\ref{fig:6} use the stricter cross-signer protocols; allowing same-signer matches raises whole-population INCLUDE Hit@1 by 20.8 points for S-G (SupCon) and 13.5 points for the separated H-G (InfoNCE).
\begin{table}[htbp]
\caption{Retrieval on the ASL-Citizen test split (11 signers unseen in training) in published work and in this work: Recall@k = Hit@k (\%) and MRR. The protocols differ in what is ranked (fourth column), so the rows are comparable only approximately. This work: whole-population cross-signer protocol, mean over seeds 0--2. --: not reported.}\label{tab:5}%
\footnotesize\setlength{\tabcolsep}{3pt}
\begin{tabularx}{\linewidth}{@{}>{\raggedright\arraybackslash\hsize=1.017\hsize}X>{\raggedright\arraybackslash\hsize=1.017\hsize}X>{\raggedright\arraybackslash\hsize=0.678\hsize}X>{\raggedright\arraybackslash\hsize=1.288\hsize}Xcccc@{}}
\toprule
Work & Model, input & Trained on ASL-Citizen & Ranked set & R@1 & R@5 & R@10 & MRR \\
\midrule
Desai et al. \cite{ref28} & I3D classifier, RGB & yes & 2,731 classes by classifier score & 63.1 & 86.1 & 90.9 & 0.73 \\
Desai et al. \cite{ref28} & ST-GCN classifier, pose & yes & 2,731 classes by classifier score & 59.5 & 82.7 & 88.1 & 0.70 \\
\addlinespace[3pt]
Jiang et al. \cite{ref18} & nDTWpp, learning-free, pose & no & 10 nearest training clips & 19 & -- & -- & -- \\
\addlinespace[3pt]
Wong et al. \cite{ref44} & SignMatch, pose & no & dictionary entries & 57.3 & 83.7 & -- & -- \\
Wong et al. \cite{ref44} & SignMatch, RGB & no & dictionary entries & 67.1 & 91.8 & -- & -- \\
\addlinespace[3pt]
\addlinespace[2pt]
\multicolumn{8}{@{}l}{\textit{This work}} \\
\hspace{0.8em}S-G, SupCon & pose, 3.6 M & yes & clips of the other test signers & 57.3 & 82.8 & 88.7 & 0.68 \\
\addlinespace[3pt]
\addlinespace[2pt]
\multicolumn{8}{@{}l}{\textit{This work}} \\
\hspace{0.8em}S-G, InfoNCE & pose, 3.6 M & yes & clips of the other test signers & 61.6 & 85.5 & 90.6 & 0.72 \\
\addlinespace[3pt]
\addlinespace[2pt]
\multicolumn{8}{@{}l}{\textit{This work}} \\
\hspace{0.8em}\textbf{H-G separated, InfoNCE} & \textbf{pose, 3.6 M} & \textbf{yes} & \textbf{clips of the other test signers} & \textbf{65.4} & \textbf{87.8} & \textbf{92.3} & \textbf{0.75} \\
\botrule
\end{tabularx}
\end{table}
Table~\ref{tab:5} places the in-language results next to the published ASL-Citizen retrieval figures. The benchmark paper ranks the 2,731 classes by the score of a classifier trained on ASL-Citizen \cite{ref28}; SignMatch ranks dictionary entries with an encoder trained on continuous British and American signing but not on ASL-Citizen \cite{ref44}; nDTWpp is learning-free \cite{ref18}; and the encoders of this work rank the clips of the other test signers. With that caveat, the reference S-G encoder (57.3 \% Recall@1) is at the level of the benchmark's pose classifier, and the best configuration (65.4 \%) is above both benchmark classifiers and above the pose variant of SignMatch, while remaining below the RGB variant of SignMatch. The encoders of this work were not designed for in-language accuracy and are much smaller than the RGB models; the comparison shows that the configurations that transfer best are also the strongest on the source language.

\subsection{Dictionary search in Indian Sign Language}
\begin{table}[htbp]
\caption{Dictionary search on CISLR (Indian Sign Language): 2,285 queries against one entry for each of 4,765 words; Hit@k (\%). This work: mean $\pm$ SD over seeds. I3D: published single result (CISLR\_v1.0-a).}\label{tab:6}%
\footnotesize\setlength{\tabcolsep}{3pt}
\begin{tabularx}{\linewidth}{@{}>{\raggedright\arraybackslash\hsize=1.564\hsize}X>{\raggedright\arraybackslash\hsize=0.436\hsize}Xcccc@{}}
\toprule
System & Input & Seeds & Hit@1 & Hit@5 & Hit@10 \\
\midrule
I3D, WLASL \cite{ref3}, published & RGB & 1 & 16.8 & 20.0 & 22.6 \\
\addlinespace[3pt]
Raw keypoints, all landmarks & pose & -- & 11.9 & 15.8 & 18.1 \\
\addlinespace[3pt]
Raw keypoints, hands & pose & -- & 12.3 & 16.5 & 18.8 \\
\addlinespace[3pt]
Untrained encoder & pose & 1 & 13.0 & 16.2 & 17.9 \\
\addlinespace[3pt]
\addlinespace[2pt]
\multicolumn{6}{@{}l}{\textit{This work}} \\
\hspace{0.8em}S-G, SupCon & pose & 8 & 17.3 $\pm$ 0.2 & 23.0 $\pm$ 0.4 & 26.0 $\pm$ 0.6 \\
\addlinespace[3pt]
\addlinespace[2pt]
\multicolumn{6}{@{}l}{\textit{This work}} \\
\hspace{0.8em}S-G, InfoNCE & pose & 8 & 18.4 $\pm$ 0.3 & 24.4 $\pm$ 0.6 & 27.3 $\pm$ 0.5 \\
\addlinespace[3pt]
\addlinespace[2pt]
\multicolumn{6}{@{}l}{\textit{This work}} \\
\hspace{0.8em}S-G separated, SupCon & pose & 3 & 18.2 $\pm$ 0.1 & 25.5 $\pm$ 0.2 & 28.9 $\pm$ 0.4 \\
\addlinespace[3pt]
\addlinespace[2pt]
\multicolumn{6}{@{}l}{\textit{This work}} \\
\hspace{0.8em}H-G separated, InfoNCE & pose & 3 & 19.0 $\pm$ 0.4 & 25.7 $\pm$ 0.2 & 28.6 $\pm$ 0.4 \\
\botrule
\end{tabularx}
\end{table}
The six targets simulate a dictionary by enrolling clips of research corpora; Table~\ref{tab:6} reports search over the real dictionary entries of CISLR (Section 4) under its one-shot protocol (Section 6). Every seed of every trained configuration finds the correct word first more often than the WLASL-trained I3D features of the original paper (lowest seed 17.1 \% against 16.8 \%), from keypoints alone, and the difference grows to 3.4--6.3 points at Hit@10. Training on ASL adds 4.3--7.2 points over the untrained encoder and raw keypoints, and the configurations keep the order of the six-target ablation: InfoNCE above SupCon, and the separated above the joint encoder under SupCon. Absolute accuracy remains low: even the best configuration misses the word at rank one for about 81 \% of queries among 4,765 words. The comparison with I3D is between available encoders with different inputs (RGB against keypoints) and source corpora (WLASL against ASL-Citizen), not between architectures. Leaving out the videos with faulty headers raises Hit@1 by 0.7--1.3 points for every model, so their inclusion does not favour the results, and 2 of the 2,285 queries share a source video with their dictionary entry.

\section{Discussion}
\textbf{Interpretation.} The results answer the three open questions of Section 1. Transfer from ASL to unrelated sign languages is substantial and consistent across six languages, and a learned encoder clearly beats learning-free matching even with one enrolled clip. Among the design choices, the objective and the separation of articulators matter and the geometry matters little. The separation result suggests that cross-part attention lets the body, which is visible in every frame, dominate representations that should be driven by the hands; isolating each articulator and pooling afterwards keeps the hand information that transfers across languages. This agrees with SignSeek, which encodes each articulator with its own graph network before temporal fusion \cite{ref12}; the present ablation isolates this choice and shows that it raises cross-lingual accuracy for the heads that pool the parts into one embedding. The geometry result is a useful negative: with matched controls and eight seeds, replacing the Euclidean space by a hyperbolic one changes accuracy by $-$2.0 to +1.1 points depending on the head and objective, so gains reported for hyperbolic sign models should be checked against norm-matched Euclidean controls.
\textbf{Model size and retrieval time.} The encoder has 3.47 M parameters and the global head 0.10 M; a dictionary entry is one 128-dimensional vector (512 bytes), so the gallery of a 10,000-sign dictionary occupies about 5 MB. Retrieval was timed on a CPU without a GPU (Intel Xeon Gold 6426Y), with one query at a time. Turning the landmarks of a clip into an embedding takes 14--16 ms on one thread and 6.6--6.9 ms on four, for both the joint S-G and the separated H-G encoder. Comparing the query with a 10,000-entry gallery and returning the ten best matches takes 0.18 ms with the spherical head and 0.70 ms with the hyperbolic one, whose distance is more expensive; with 100,000 entries, 3.0 and 17 ms. The dominant cost is landmark extraction: MediaPipe Holistic needs about 30 ms per 1080p frame on the same CPU (about 33 frames per second), so landmarks can be extracted while the sign is performed in front of a camera, and the answer follows within a few tens of milliseconds after the sign ends. Adding signs or a new sign language requires only embedding their example videos, which needs no training and no GPU.
\textbf{Limitations.} Only one source language is studied. Inputs are 2-D body and hand landmarks without face, so non-manual features are not modelled. WLPSL and SSL lack signer identities and are signer-uncontrolled. On the six corpora the low-resource setting is simulated by enrolling one or four clips per sign in corpora that are not themselves low-resource. CISLR uses real dictionary entries, but its queries are dictionary videos rather than recordings by people searching the dictionary, and it has no signer identities, so signer overlap between entry and query cannot be excluded there. Language relatedness (French Sign Language influenced both ASL and Russian Sign Language) may favour some targets. INCLUDE and LSA64 were used in the authors' earlier work, and seed-0 pilot results on three targets were seen before the design was extended. The part separation, cross-entropy, OpenHands and eight-seed experiments were added after the first three-seed results. Velocities are computed after resampling to the fixed clip length, so in clips shorter than that length (the 10 fps SSL recordings most of all) repeated frames carry zero velocity. Hyperparameters were not tuned per objective or geometry. Two Kaggle datasets have no documented licence and are used for evaluation only.

\section{Conclusions}
A pose encoder trained only on American Sign Language retrieves signs of six unseen sign languages far better than learning-free keypoint matching and better than available ASL-trained graph networks, with a single enrolled example per sign, and on real Indian Sign Language dictionary entries (CISLR) it exceeds the published RGB baseline. Training it with InfoNCE and encoding body and hands separately before pooling are the two choices that matter, together raising one-shot accuracy from 59.0 to 68.0 \%; the choice of spherical, Euclidean or hyperbolic embedding space is a minor factor. Over eight seeds with the joint encoder, spherical, Euclidean and hyperbolic heads differ by at most 1.6 points, except the hyperbolic token-attention head with InfoNCE, which is 1.4--2.0 points below the other two (2.9--3.9 points with the separated encoder, three seeds). The cross-signer protocols, saved trials and code will be released so that encoders for low-resource sign languages can be compared on equal terms. Future work includes other source languages, face landmarks and harder negative sampling.
\textbf{Data and code availability.} Code, trial files and dataset manifests will be released on publication. The full set of result tables is provided as supplementary material.

\backmatter
\bmhead{Supplementary information}
The supplementary material contains the per-target result tables, the dimension sweep, the significance tests of the further ablations, the occlusion analysis and the dictionary-search details referred to in the text.
\section*{Declarations}
\begin{itemize}
\item Funding: This work was funded by the Variable Energy Cyclotron Centre, Kolkata, India.
\item Conflict of interest: The authors declare no competing interests.
\item Ethics approval and consent to participate: Not applicable; only publicly released datasets were used.
\item Consent for publication: Not applicable.
\item Data availability: All datasets used are publicly available from the sources cited in Table~\ref{tab:1}.
\item Code availability: Code, trial files and dataset manifests will be released on publication.
\item Author contribution: All authors contributed equally to this work.
\end{itemize}


\begin{thebibliography}{99}
\bibitem{ref1} United Nations, ``International Day of Sign Languages, 23 September.'' [Online]. Available: https://www.un.org/en/observances/sign-languages-day (accessed Nov. 12, 2026).
\bibitem{ref2} Indian Sign Language Research and Training Centre (ISLRTC), \textit{Indian Sign Language Dictionary}, 3rd ed. New Delhi, India: ISLRTC, Dept. of Empowerment of Persons with Disabilities, Ministry of Social Justice and Empowerment, 2021. [Online]. Available: https://islrtc.nic.in/
\bibitem{ref3} A. Joshi \textit{et al.}, ``CISLR: Corpus for Indian Sign Language recognition,'' in \textit{Proc. 2022 Conference on Empirical Methods in Natural Language Processing}, Association for Computational Linguistics, 2022, pp. 10357--10366, doi: 10.18653/v1/2022.emnlp-main.707.
\bibitem{ref4} A. Sridhar, R. G. Ganesan, P. Kumar, and M. Khapra, ``INCLUDE: A large scale dataset for Indian sign language recognition,'' in \textit{Proc. 28th ACM International Conference on Multimedia}, 2020, pp. 1366--1375, doi: 10.1145/3394171.3413528.
\bibitem{ref5} M. De Coster and J. Dambre, ``Querying a sign language dictionary with videos using dense vector search,'' in \textit{2023 IEEE International Conference on Acoustics, Speech, and Signal Processing Workshops (ICASSPW), 8th Workshop on Sign Language Translation and Avatar Technology (SLTAT)}, IEEE, 2023, pp. 1--5, doi: 10.1109/ICASSPW59220.2023.10193531.
\bibitem{ref6} T. Vandendriessche, C. Brosens, H. De Durpel, M. De Coster, and J. Dambre, ``SignBuddy: From sign language research to scalable co-created solutions,'' \textit{Universal Access in the Information Society}, vol. 25, Art. no. 45, 2026, doi: 10.1007/s10209-026-01310-1.
\bibitem{ref7} W. C. Stokoe, \textit{Sign Language Structure: An Outline of the Visual Communication Systems of the American Deaf} (Studies in Linguistics, Occasional Papers, no. 8). Buffalo, NY, USA: Univ. of Buffalo, 1960.
\bibitem{ref8} P. Selvaraj, G. NC, P. Kumar, and M. Khapra, ``OpenHands: Making sign language recognition accessible with pose-based pretrained models across languages,'' in \textit{Proc. 60th Annual Meeting of the Association for Computational Linguistics (Volume 1: Long Papers)}, 2022, pp. 2114--2133, doi: 10.18653/v1/2022.acl-long.150.
\bibitem{ref9} M. De Coster, E. Rushe, R. Holmes, A. Ventresque, and J. Dambre, ``Towards the extraction of robust sign embeddings for low resource sign language recognition,'' arXiv:2306.17558, 2023.
\bibitem{ref10} G. NC, M. Ladi, S. Negi, P. Selvaraj, P. Kumar, and M. Khapra, ``Addressing resource scarcity across sign languages with multilingual pretraining and unified-vocabulary datasets,'' in \textit{Advances in Neural Information Processing Systems 35 (NeurIPS 2022), Datasets and Benchmarks Track}, 2022.
\bibitem{ref11} Y. C. Bilge, N. Ikizler-Cinbis, and R. G. Cinbis, ``Cross-lingual few-shot sign language recognition,'' \textit{Pattern Recognition}, vol. 151, Art. no. 110374, 2024, doi: 10.1016/j.patcog.2024.110374.
\bibitem{ref12} S. Asasi, O. Mercanoglu Sincan, and R. Bowden, ``SignSeek: Learning transferable representations for sign dictionary retrieval,'' arXiv:2609.03695, 2026.
\bibitem{ref13} U. Sarkar, B. Chattopadhyay, T. Samanta, H. K. Pandey, and N. P. Yalagoud, ``Pose-based contrastive representation learning for sign languages,'' in \textit{2026 International Conference on Computing, Electronics \& Communications Engineering (ICCECE)}, IEEE, 2026, doi: 10.1109/ICCECE69715.2026.11493072.
\bibitem{ref14} B. Orazumbekov, D. Bayanov, A. Kaltay, and A. Sandygulova, ``A video-based reverse dictionary for sign language using gesture similarity,'' in \textit{Proc. 12th Workshop on the Representation and Processing of Sign Languages}, ELRA, 2026, pp. 398--407.
\bibitem{ref15} H. Sakoe and S. Chiba, ``Dynamic programming algorithm optimization for spoken word recognition,'' \textit{IEEE Transactions on Acoustics, Speech, and Signal Processing}, vol. 26, no. 1, pp. 43--49, 1978.
\bibitem{ref16} H. Wang, A. Stefan, S. Moradi, V. Athitsos, C. Neidle, and F. Kamangar, ``A system for large vocabulary sign search,'' in \textit{Trends and Topics in Computer Vision (ECCV 2010 Workshops)}, LNCS 6553, Springer, 2012, pp. 342--353.
\bibitem{ref17} Y. Dong, ``Angle based DTW for large vocabulary sign retrieval,'' in \textit{Advances in Visual Computing (ISVC 2025)}, LNCS 16397, Springer, 2026, pp. 389--403, doi: 10.1007/978-3-032-14495-9\_30.
\bibitem{ref18} Z. Jiang \textit{et al.}, ``Meaningful pose-based sign language evaluation,'' in \textit{Proc. Tenth Conference on Machine Translation (WMT 2025)}, 2025, pp. 64--80, doi: 10.18653/v1/2025.wmt-1.4.
\bibitem{ref19} T. Jiang, N. C. Camg\"{o}z, and R. Bowden, ``Looking for the signs: Identifying isolated sign instances in continuous video footage,'' in \textit{2021 16th IEEE International Conference on Automatic Face and Gesture Recognition (FG 2021)}, IEEE, 2021, doi: 10.1109/FG52635.2021.9667037.
\bibitem{ref20} M. Nickel and D. Kiela, ``Poincar\'{e} embeddings for learning hierarchical representations,'' in \textit{Proc. Adv. Neural Inf. Process. Syst. (NeurIPS)}, 2017.
\bibitem{ref21} O. Ganea, G. B\'{e}cigneul, and T. Hofmann, ``Hyperbolic neural networks,'' in \textit{Proc. Adv. Neural Inf. Process. Syst. (NeurIPS)}, 2018.
\bibitem{ref22} A. Ermolov, L. Mirvakhabova, V. Khrulkov, N. Sebe, and I. Oseledets, ``Hyperbolic vision transformers: Combining improvements in metric learning,'' in \textit{Proc. IEEE/CVF Conf. Comput. Vis. Pattern Recognit. (CVPR)}, 2022.
\bibitem{ref23} E. Fish and R. Bowden, ``Geo-Sign: Hyperbolic contrastive regularisation for geometrically aware sign language translation,'' in \textit{Advances in Neural Information Processing Systems (NeurIPS)}, 2025, arXiv:2506.00129.
\bibitem{ref24} M. Pu, M. K. Lim, C. Y. Chong, and C. C. Loy, ``LA-Sign: Looped Transformers with geometry-aware alignment for skeleton-based sign language recognition,'' arXiv:2603.29057, 2026.
\bibitem{ref25} P. Khosla \textit{et al.}, ``Supervised contrastive learning,'' in \textit{Proc. Adv. Neural Inf. Process. Syst. (NeurIPS)}, 2020.
\bibitem{ref26} A. van den Oord, Y. Li, and O. Vinyals, ``Representation learning with contrastive predictive coding,'' arXiv:1807.03748, 2018.
\bibitem{ref27} E. W. Teh, T. DeVries, and G. W. Taylor, ``ProxyNCA++: Revisiting and revitalizing proxy neighborhood component analysis,'' in \textit{Proc. Eur. Conf. Comput. Vis. (ECCV)}, 2020.
\bibitem{ref28} A. Desai \textit{et al.}, ``ASL Citizen: A community-sourced dataset for advancing isolated sign language recognition,'' in \textit{Advances in Neural Information Processing Systems 36 (NeurIPS 2023), Datasets and Benchmarks Track}, 2023.
\bibitem{ref29} O. M. Sincan and H. Y. Keles, ``AUTSL: A large scale multi-modal Turkish sign language dataset and baseline methods,'' \textit{IEEE Access}, vol. 8, pp. 181340--181355, 2020, doi: 10.1109/ACCESS.2020.3028072.
\bibitem{ref30} A. Kapitanov, K. Kvanchiani, A. Nagaev, and E. Petrova, ``Slovo: Russian Sign Language dataset,'' in \textit{Computer Vision Systems (ICVS 2023)}, LNCS, Springer, 2023, pp. 63--73, doi: 10.1007/978-3-031-44137-0\_6.
\bibitem{ref31} F. Ronchetti, F. Quiroga, C. Estrebou, L. Lanzarini, and A. Rosete, ``LSA64: An Argentinian sign language dataset,'' in \textit{XXII Congreso Argentino de Ciencias de la Computaci\'{o}n (CACIC 2016)}, 2016, pp. 794--803.
\bibitem{ref32} J. Naeem, Z. U. A. Asif, and A. U. Rahman, ``Word-level Pakistani Sign Language recognition,'' in \textit{Computational Intelligent Systems: Applications of AI and Machine Learning} (Infosys Science Foundation Series), M. I. Uddin, F. Aziz, and H. Jafari, Eds. Singapore: Springer, 2026, pp. 187--200, doi: 10.1007/978-981-95-2212-5\_13.
\bibitem{ref33} Y. Abhishek and T. G. D. K. Sumanathilaka, ``SSL400: A comprehensive word level dataset for Sinhala Sign Language recognition,'' in \textit{2025 5th International Conference on Advanced Research in Computing (ICARC)}, IEEE, 2025, pp. 1--6, doi: 10.1109/ICARC64760.2025.10962991.
\bibitem{ref34} C. Lugaresi \textit{et al.}, ``MediaPipe: A framework for building perception pipelines,'' arXiv:1906.08172, 2019.
\bibitem{ref35} K. Musgrave, S. Belongie, and S.-N. Lim, ``A metric learning reality check,'' in \textit{Proc. Eur. Conf. Comput. Vis. (ECCV)}, 2020.
\bibitem{ref36} S. Jiang, B. Sun, L. Wang, Y. Bai, K. Li, and Y. Fu, ``Skeleton aware multi-modal sign language recognition,'' in \textit{Proc. IEEE/CVF Conference on Computer Vision and Pattern Recognition Workshops (CVPRW)}, 2021, pp. 3408--3418, doi: 10.1109/CVPRW53098.2021.00380.
\bibitem{ref37} S. Yan, Y. Xiong, and D. Lin, ``Spatial temporal graph convolutional networks for skeleton-based action recognition,'' in \textit{Proc. AAAI Conf. Artif. Intell.}, vol. 32, no. 1, 2018, pp. 7444--7452, doi: 10.1609/aaai.v32i1.12328.
\bibitem{ref38} D. Li, C. Rodriguez Opazo, X. Yu, and H. Li, ``Word-level deep sign language recognition from video: A new large-scale dataset and methods comparison,'' in \textit{Proc. IEEE/CVF Winter Conference on Applications of Computer Vision (WACV)}, 2020, pp. 1448--1458, doi: 10.1109/WACV45572.2020.9093512.
\bibitem{ref39} A. Vaswani \textit{et al.}, ``Attention is all you need,'' in \textit{Proc. Adv. Neural Inf. Process. Syst. (NeurIPS)}, 2017.
\bibitem{ref40} Y. Guo, X. Wang, Y. Chen, and S. X. Yu, ``Clipped hyperbolic classifiers are super-hyperbolic classifiers,'' in \textit{Proc. IEEE/CVF Conf. Comput. Vis. Pattern Recognit. (CVPR)}, 2022.
\bibitem{ref41} M. Fr\'{e}chet, ``Les \'{e}l\'{e}ments al\'{e}atoires de nature quelconque dans un espace distanci\'{e},'' \textit{Annales de l'Institut Henri Poincar\'{e}}, vol. 10, no. 4, pp. 215--310, 1948.
\bibitem{ref42} H. Karcher, ``Riemannian center of mass and mollifier smoothing,'' \textit{Communications on Pure and Applied Mathematics}, vol. 30, no. 5, pp. 509--541, 1977, doi: 10.1002/cpa.3160300502.
\bibitem{ref43} I. Loshchilov and F. Hutter, ``Decoupled weight decay regularization,'' in \textit{Proc. Int. Conf. Learn. Represent. (ICLR)}, 2019.
\bibitem{ref44} R. Wong, Y. Jang, L. Momeni, G. Varol, and A. Zisserman, ``SignMatch: Matching dictionary signs to continuous sign language video,'' in \textit{Proc. British Machine Vision Conference (BMVC 2026)}, 2026, arXiv:2609.01886.
\end{thebibliography}
\end{document}